\documentclass{article} 
\usepackage{iclr2027_conference,times}

\usepackage{amsmath,amsfonts,bm}

\def\eqref#1{equation~\ref{#1}}

\def\1{\bm{1}}

\DeclareMathAlphabet{\mathsfit}{\encodingdefault}{\sfdefault}{m}{sl}
\SetMathAlphabet{\mathsfit}{bold}{\encodingdefault}{\sfdefault}{bx}{n}

\usepackage{hyperref}
\usepackage{url}
\usepackage[utf8]{inputenc} 
\usepackage[T1]{fontenc}    
\usepackage{booktabs}       
\usepackage{amsfonts}       
\usepackage{nicefrac}       
\usepackage{microtype}      
\usepackage{xcolor} 
\usepackage{xspace}
\usepackage{wrapfig}
\usepackage{makecell}
\usepackage{graphicx}
\usepackage{multirow}
\usepackage{pgfplots}
\pgfplotsset{compat=1.18}

\usepackage{subcaption}

\usepackage{tabularx}
\usepackage{placeins}
\usepackage{array}
\usepackage{float}
\usepackage[most]{tcolorbox}

\usepackage{amsmath}
\usepackage{amssymb}
\usepackage{mathtools}
\usepackage{amsthm}

\newcommand{\ada}{\textsc{AdaptArena}\xspace}
\newcommand{\ouragentname}{\textsc{AdaptiveAgent}\xspace}

\title{\ada: Evaluating Test-Time Personalization of Web Agents}

\iclrfinalcopy 

\newcommand{\up}[1]{\textnormal{\,\textsuperscript{#1}}}

\newcommand{\upns}[1]{\textnormal{\textsuperscript{#1}}}

\def\mila{1}
\def\mcgill{2}
\def\service{3}
\def\cifar{4}
\def\laval{5}

\author{
\textbf{Dongchan Shin}\up{\mila}
\textbf{Xing Han Lù}\up{\mila,\kern-0.03em\mcgill}
\textbf{Jiaqi Deng}\up{\mila}
\textbf{Jay Gala}\up{\mila,\kern-0.03em\mcgill}
\textbf{Tomás Vergara Browne}\up{\mila,\kern-0.03em\mcgill} \\
\textbf{Jaewon Moon}\up{\mcgill}
\textbf{Fengyuan Liu}\up{\mila,\kern-0.03em\mcgill}
\textbf{Alexandre Drouin}\up{\mila,\kern-0.03em\service,\kern-0.03em\laval}
\textbf{Siva Reddy}\up{*\kern-0.03em\mila,\kern-0.03em\mcgill,\kern-0.03em\service,\kern-0.03em\cifar}
\textbf{Alexandre Lacoste}\up{*\kern-0.03em\service}
\\
\upns{\mila}Mila - Quebec AI Institute~~
\upns{\mcgill}McGill University~~ 
\upns{\service}ServiceNow Research~~ \\
\upns{\cifar}Canada CIFAR AI Chair~~
\upns{\laval}Université Laval~~
\upns{*}Equal advising
}

\begin{document}

\maketitle

\begin{abstract}
\label{sec:abstract}
Large language model (LLM) agents have demonstrated strong performance on complex web navigation tasks, yet they remain brittle in real-world settings where user intentions are underspecified and preferences are heterogeneous. In practice, users rarely provide explicit profiles, requiring agents to infer latent preferences from implicit signals. Despite its importance for deployment, this problem setting is largely underexplored in existing benchmarks. To address this gap, we introduce \ada, a benchmark for evaluating test-time personalization of web agents via implicit preference inference. \ada consists of 480 tasks, featuring both single-preference and double-preference scenarios. Each evaluation task must be solved by retrieving and leveraging the most relevant historical user trajectory that implicitly encodes the target preference. In addition, we introduce \ouragentname, a retrieval-based framework for standardized evaluation of implicit preference inference. Experiments reveal a substantial performance gap: while oracle agents with access to ground-truth preferences achieve an 82.92\% success rate, the evaluated LLM agents using our framework reach at most 15.62\%. Furthermore, we find that correctly inferring user preferences is necessary but not sufficient for task success, as execution failures in downstream web interactions remain a significant bottleneck even when agents align with the target preference. These findings highlight implicit preference inference and robust action grounding as key challenges for deploying reliable, user-facing web agents. We release our code: https://github.com/McGill-NLP/web-agents-test-time-adaptations
\end{abstract}

\section{Introduction}
\label{sec:intro}
Large language model (LLM) agents have recently demonstrated strong performance across a wide range of interactive tasks, including web navigation ~\citep{deng2023mind2web, zhou2023webarena, he2024webvoyager, lu2024weblinx}, computer-use ~\citep{xie2024osworld, wang2025opencua} and tool-calling ~\citep{yao2024tau, wang2024gta}. In particular, web-based LLM agents have emerged as a compelling frontier for human-computer interaction. By operating directly within real-world browser environments, these agents enable end-to-end task automation and offer a practical pathway toward general-purpose autonomy. 

Despite their impressive general capabilities, current web agents struggle to adapt when deployed in novel environments. Existing research has tried to address this challenge through self-improvement mechanisms, such as learning from past experiences or environmental exploration \citep{wang2024agent, wang2025inducing, zheng2025skillweaver, Nekoei2025JustintimeEF}. While these efforts have significantly improved the robustness of web agents against new websites, a critical dimension of adaptation remains largely unexplored: the behavioral diversity of users who interact with those websites.

This gap reveals a fundamental barrier to real-world applicability. In practice, web agents must support users with diverse preferences and backgrounds, often without access to explicit profiles. Addressing this challenge through full model retraining is both costly and impractical, as it requires repeated parametric updates ~\citep{quinonero2009dataset} whenever user distributions shift. Instead, web agents should adapt to user-specific behaviors under a fixed model (i.e., without gradient-based updates). This motivates test-time personalization ~\citep{zhang2025personaagent, qu2025t} -- a form of test-time adaptation ~\citep{wang2020tent} where a deployed agent adjusts its behavior at inference time to latent user preferences without modifying model parameters. A natural paradigm for test-time personalization is to leverage a user's past interaction history, which provides implicit evidence of their preferences. This setting aligns with real-world usage, where explicit preference signals are rarely available but rich behavioral traces accumulate over time.

A few recent studies have tried to evaluate the personalization of web agents. PersonalWAB ~\citep{cai2025large} relies on explicit, text-based user profiles, while CUPID~\citep{kim2025cupid} depends on direct verbal feedback to identify user needs. However, these benchmarks operate in unrealistic settings, as such curated metadata and explicit personas are rarely available in real-world browsing scenarios. Persona2Web~\citep{kim2026persona2web} studies personalized web agents that resolve ambiguous web queries using browsing histories and user context, but primarily focuses on query disambiguation from static user histories rather than evaluating whether agents can retrieve and operationalize latent task-relevant preferences from historical interaction trajectories at test time.

\textbf{Contributions.} We introduce \ada, a benchmark for evaluating test-time personalization of web agents through implicit preference inference from historical user interaction trajectories. \ada comprises 480 tasks across three web environments, covering both single- and double-preference settings under underspecified user instructions. Unlike prior personalization benchmarks that provide explicit profiles or direct preference feedback, \ada requires agents to recover task-relevant preferences from behavioral history. We further introduce \ouragentname, a simple and reproducible retrieval-based framework that provides a standardized way to retrieve relevant historical trajectories, infer latent preferences, and incorporate them into downstream web-agent execution. Across six backbone models, \ouragentname achieves at most 15.62\% task success, compared with up to 82.92\% with ground-truth preferences, revealing a substantial gap in implicit preference inference. Our analysis further shows a consistent drop from single to double preferences across models, highlighting the difficulty of compositional personalization. Finally, we find that correct preference inference does not reliably translate into successful execution, with downstream action failures remaining common even under correct preference alignment.

\begin{figure*}[t]
    \centering
        \includegraphics[width=.95\textwidth]{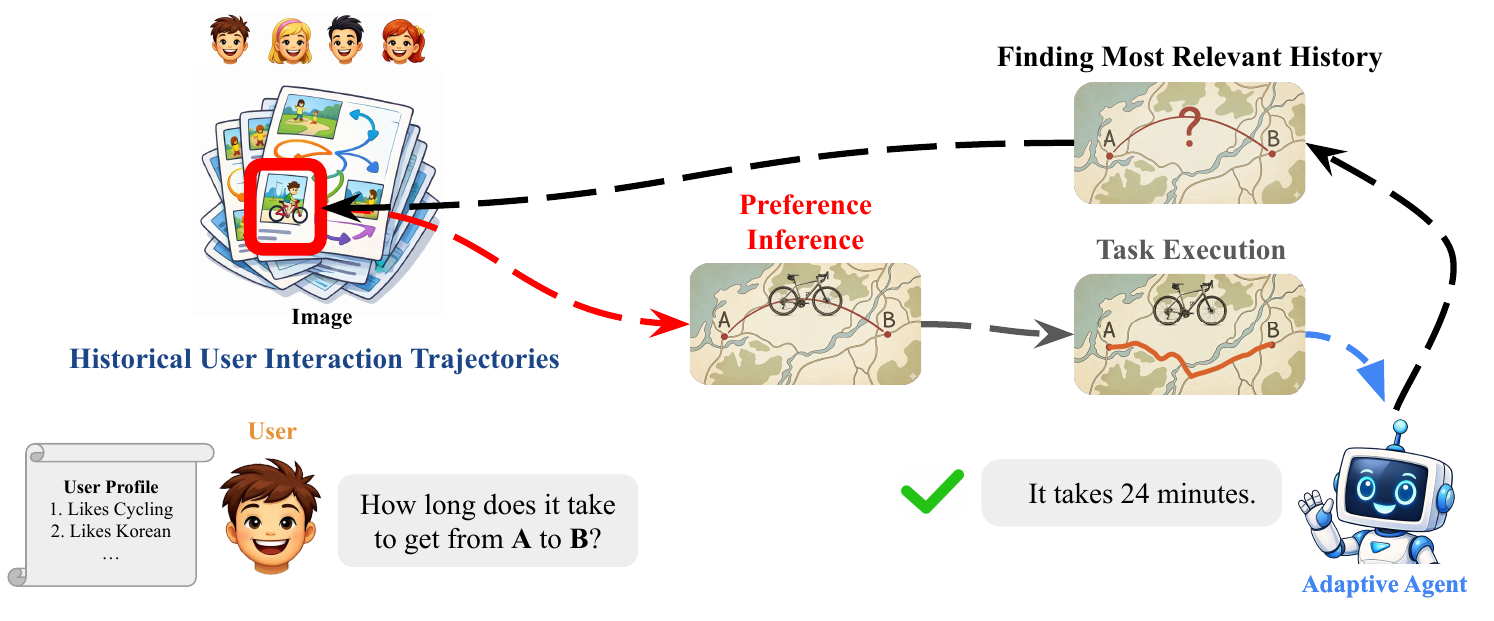}
    
    \caption{Benchmark overview. The agent resolves underspecified user tasks by inferring latent preferences from user history; in this example, the \ouragentname infers a preference for cycling and estimates the expected travel time from location A to B accordingly.}
    \label{fig:overview}
\end{figure*}

\begin{figure*}[t]
    \centering
    \scalebox{1}[1]{%
        \includegraphics[width=.95\textwidth]{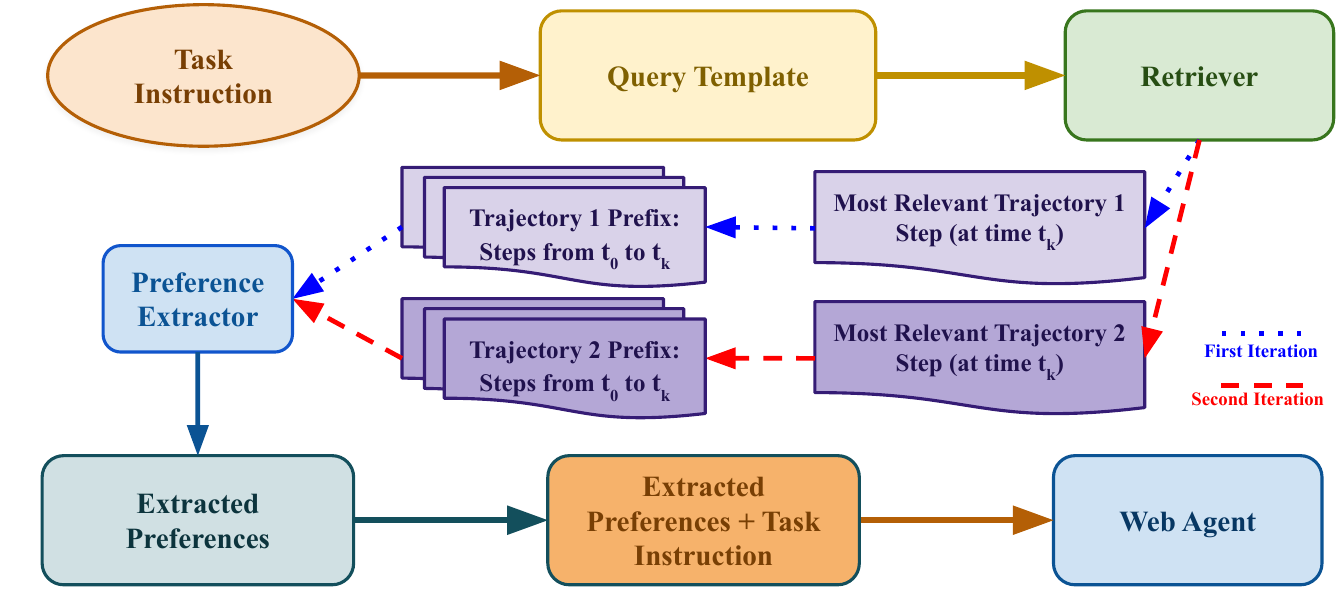}
    }
    \caption{\ouragentname Structure. The task instruction is encoded as a retrieval query to identify the single most relevant trajectory step at time $t_k$. The corresponding trajectory prefix from $t_0$ to $t_k$ is reconstructed, and the selected trajectory is then removed from the candidate pool. This process is repeated once to retrieve one additional relevant trajectory, yielding a total of two trajectories. Prefixes from the selected trajectories are aggregated into a unified sequence of screenshots, which is used to infer the user's implicit preference. The inferred preference is combined with the task instruction and provided to the web agent for execution.}
    \label{fig:adaptive-agent}
\end{figure*}

\section{Benchmark}
\label{sec:benchmark}

\subsection{Evaluation Paradigm}
\label{sec:benchmark_overview}
In an ideal evaluation setting, a personalized agent would interact with a user over an extended period,
gradually learning the user's preferences through natural interactions, and would subsequently be evaluated on new tasks that require integrating these preferences at test time.
While realistic, such an interactive protocol is impractical for benchmarking: it is costly, difficult to scale, and introduces significant reproducibility challenges due to variability in interaction length and content.

To address these limitations, we adopt a two-phase evaluation paradigm that decouples preference acquisition from test-time evaluation. In the first phase, we construct \emph{encoded user histories} by collecting interaction trajectories generated under explicit preference conditions.
In the second phase, agents are evaluated on new tasks in which all explicit preference information is removed.
As a result, agents must infer the relevant preferences solely from the provided user history and act accordingly.

\subsection{Web Environments}
We construct tasks across three web environments drawn from WebArena ~\citep{zhou2023webarena}: an e-commerce store, a Reddit discussion forum, and OpenStreetMap. These environments are selected because they represent realistic settings in which personalization and user-specific preferences naturally arise. To enable agents to interact with our web environments, we use the BrowserGym ~\citep{chezelles2025browsergym} platform, which provides accessibility trees, screenshots, and prompt interfaces. BrowserGym parses and executes actions produced by web agents, allowing for consistent and controlled interaction across tasks.

\subsection{Task Annotation}

\subsubsection{Trajectory Collection}
The primary objective of the trajectory collection phase is to construct user histories that encode clear preference signals. We define ten user personas; each persona is characterized by the same set of twelve preference categories, which are grounded across the three web environments.

Trajectory collection annotations consist of two components: a \emph{Persona Trait} and a \emph{General Instruction}. The Persona Trait encodes an explicit user preference (e.g., \emph{prefers to drive}), while the General Instruction specifies the preference-agnostic task goal (e.g., \emph{How long would it take to get from CMU to University of Pittsburgh?}). Together, these components form a user-specific instruction: \emph{I prefer driving. How long would it take to get from CMU to University of Pittsburgh?} 
For each task, we design a user intent that encodes exactly one preference, with annotations manually reviewed to ensure solvability based on that preference alone.

Trajectory collection is performed by an agent acting as a user behavior generator, instantiated using Qwen-3-VL-32B-Thinking ~\citep{bai2025qwen3vltechnicalreport}. In total, we construct 120 tasks for trajectory collection and obtain 120 corresponding user-specific trajectories, with twelve preference-aligned tasks and trajectories per
persona across ten personas. Sample trajectories and user personas are provided in the Appendix \ref{sec:benchmark}. We additionally analyze the quality and diversity of the collected
trajectories. Across the 120 trajectories, the action distribution is
dominated by clicks and fills, reflecting the interaction
repertoire required by the underlying environments. The full action-type and trajectory-length
statistics are reported in Appendix~\ref{subsec:traj}. To verify that the preference signal encoded in trajectories is recoverable
without access to explicit annotations, we conduct an independent human
study. Two annotators unaffiliated with the project are shown 60 shuffled
raw trajectories with preference information removed and asked to identify
the underlying preference. Humans recover the gold preference in
$85.83\%$ of cases on average, substantially above the
$51.67\%$ coverage obtained by our reconstructed-profile baseline.
\FloatBarrier

\FloatBarrier
\subsubsection{Deployment}
This phase constitutes the primary evaluation stage of our benchmark, where agents are assessed on their ability to leverage user history at test time. To isolate this capability, we exclude any explicit preference information from the user intents. Agents must instead infer the appropriate preference from the provided user history and complete the task accordingly. Each task is designed such that successful completion depends on inferring a single, well-defined user preference. All deployment tasks undergo manual review to verify that (1) the task is underspecified without access to user history, and (2) it becomes unambiguous and solvable once the corresponding preference is inferred from the history. The collection is category-balanced by construction, with ten personas
covering the same twelve preference categories. For deployment, we construct
480 tasks with equal numbers of
single- and double-preference tasks. Task distributions are shown in \ref{tab:task_dist}.

\subsection{Evaluation Metrics}

\subsubsection{Task Success Rate}
Following prior work \citep{zhou2023webarena}, we measure task performance using a \emph{functional automatic evaluator} that returns a binary success label for each task. For every intent, a reference object is constructed to specify the success criteria. We employ two types of reference objects. The first is a reference answer, evaluated via string matching against the agent's output. The second is a \emph{reference program}, used for tasks requiring information extraction. Each reference program specifies a reference URL, JavaScript-based locators for identifying target elements, and the expected content (a string or list), which is matched against the extracted result using exact or partial matching rules.

\subsubsection{Preference Alignment Score}
We evaluate an agent's ability to infer user preferences using a \textit{Preference Alignment Score} computed by an LLM-based judge ~\citep{gu2024survey} implemented with Gemini-3-Flash. For each episode, the judge is provided with the complete interaction trajectory, along with a description of the target user preference. Based on the agent's observed navigation and interaction patterns, the judge determines whether the trajectory is consistent with the specified preference. To validate the reliability of this evaluation, we assess agreement between the LLM judge and independent human annotations using Cohen's $\kappa$~\citep{cohen1960coefficient}.

\section{Experiment}

\begin{table*}[t]
\centering
\caption{Task success rates (\%) of different models under varying levels of preference availability across 480 tasks. \textit{Reconstructed Profile} benefits from partial access to ground-truth user preferences. \textit{User-Centric} and \textit{Oracle} assume full access to ground-truth user information and therefore serve as upper bounds, reflecting idealized conditions beyond realistic deployment scenarios.}
\label{tab:main_result}
\resizebox{\textwidth}{!}{%
\begin{tabular}{lccccc}
\toprule

Model & 
\multicolumn{2}{c}{Baseline} & 
\multicolumn{1}{c}{Our Method} & 
\multicolumn{2}{c}{Upper Bound} \\

\cmidrule(lr){2-3} \cmidrule(lr){4-4} \cmidrule(lr){5-6}

 & \makecell{No \\ Profile} & \makecell{Reconstructed \\ Profile} & \ouragentname & \makecell{User-Centric} & \makecell{Oracle} \\
\midrule

Gemini-3-Pro & 7.71\% & 17.71\% & 15.62\% & 68.33\% & 82.92\% \\
Gemini-3-Flash & 9.17\% & 17.71\% & 15.00\% & 56.04\% & 72.29\% \\
GPT-5.4 & 8.96\% & 14.17\% & 12.71\% & 51.04\% & 71.04\% \\
GPT-5-mini & 7.92\% & 14.17\% & 11.67\% & 41.87\% & 62.92\% \\
Qwen-3.5-27B & 8.96\% & 15.00\% & 11.04\% & 46.46\% & 66.87\% \\
Qwen-3-VL-32B & 7.29\% & 15.42\% & 10.42\% & 25.21\% & 40.83\% \\

\bottomrule
\end{tabular}%
}
\end{table*}

\subsection{Backbone Models}
\label{sec:models}

We evaluate six backbone models for our web agents: Gemini-3-Pro, Gemini-3-Flash ~\citep{google2025gemini3}, GPT-5.4, GPT-5-mini ~\citep{openai_gpt54_2025, openai2025gpt5dev}, Qwen-3.5-27B and Qwen-3-VL-32B-Thinking ~\citep{qwen35blog, bai2025qwen3vltechnicalreport}. These models were selected for their strong reasoning capabilities and robust multimodal processing abilities, which are essential for web interactions. 

\subsection{\ouragentname}

We introduce a simple \ouragentname framework that explicitly separates preference inference from task execution, without assuming access to any gold or pre-reconstructed user profile at inference time. Instead, the framework relies solely on past interaction trajectories as its source of supervision. At inference time, the framework performs user-specific retrieval over a database of past interaction trajectories, where each trajectory is represented as an ordered sequence of screenshot-based steps $\{s_0, s_1, \dots, s_T\}$. Given a task instruction, we first encode it into a shared embedding space and iteratively retrieve the most relevant trajectory using a CLIP-based similarity search with FAISS. From each retrieved trajectory, the framework then selects the step $s_k$ (where $k \in \{0, \dots, T\}$) most relevant to the current task. After retrieval, the selected trajectory is removed from the candidate pool. This process is repeated once to retrieve one additional relevant trajectory, yielding a total of two trajectories.
For each retrieved trajectory, we construct a trajectory prefix spanning from the initial step $s_0$ to the selected step $s_k$. We then concatenate the prefixes from all selected trajectories into a single set of screenshots. The aggregated sequence of screenshots is provided to a preference extractor, implemented using GPT-5-nano, which infers the user’s latent preferences. The inferred preference is expressed as a concise natural language summary and is then incorporated into the agent’s prompt as an auxiliary context. We provide implementation details in Appendix \ref{sec:c4}.

\subsection{Alternative Agent Designs}
\label{sec:alt_designs}
\paragraph{Baselines.}
The \textbf{No Profile} baseline measures how web agents perform when no user history is provided. The agent must rely solely on the task description and the observable environment. Additionally, the \textbf{Reconstructed Profile} is a strong baseline designed to assess whether user preferences can be inferred by jointly inspecting all trajectories from a single user, without requiring an explicit retrieval mechanism. During trajectory collection, the user behavior generator is instructed with an explicit user preference, which may implicitly influence its intermediate reasoning traces and action choices. To evaluate whether such information alone is sufficient for preference identification, we reconstruct a user profile by summarizing the agent's observed thoughts and actions and applying a strong LLM (GPT-5) to infer the underlying preference. Notably, this process effectively exposes the model to partial information about the ground-truth preference distribution. Empirically, the reconstructed profiles achieve an average coverage rate of \textit{51.67\%} with respect to the gold profiles across ten personas, highlighting that a substantial portion of user preferences can be recovered from trajectories alone. The resulting  profile is then provided to the agent as a textual input during test-time. Sample reconstructed profiles are provided in the Appendix \ref{sec:b3}.

\paragraph{Upper Bounds.}
\textbf{User-Centric} agent is provided with the complete gold user profile, representing a realistic personalization setting in which multiple user preferences are available simultaneously. The agent must first identify which preference is relevant to the current task and then apply it appropriately during task execution. This setting therefore evaluates the full personalization process, including both preference identification and preference-conditioned execution, without requiring the agent to infer preferences that are absent from the provided profile. \textbf{Oracle} agent has direct access to the single gold preference required for the task. Unlike the User-Centric setting, this removes the need for preference identification and isolates the agent's ability to correctly apply a known preference.

\subsection{Results}
\label{sec:results}
\subsubsection{Task Success Rate}
\label{sec:task_success_rate}

\paragraph{No Profile and Reconstructed Profile baselines.}
Table \ref{tab:main_result} shows that in the absence of user history, all models perform poorly, with success rates ranging from 7.29\% to 9.17\%, highlighting the difficulty of resolving underspecified tasks without user-specific preference signals. Introducing reconstructed user profiles yields consistent improvements of approximately 5\% to 10\% across all models. For example, Gemini-3-Pro improves from 7.71\% to 17.71\%, and GPT-5-mini from 7.92\% to 14.17\%. It is important to note that reconstructed profiles are built with partial access to gold user preferences. As such, this setting provides a non-trivial amount of oracle information, and should be interpreted as a semi-privileged baseline rather than a fully realistic deployment scenario.

\paragraph{\ouragentname.}
In contrast, \ouragentname operates without access to any gold preference signals, instead relying solely on retrieval over past interaction trajectories at inference time. Despite this stricter and more realistic setting, our method remains competitive with the reconstructed profile baseline across all models. For instance, GPT-5.4 achieves 12.71\% (vs.\ 14.17\%) and Gemini-3-Pro achieves 15.62\% (vs.\ 17.71\%). Notably, \ouragentname consistently outperforms the no-profile baseline, demonstrating that even implicit preference signals recovered through retrieval provide meaningful gains. This result suggests that trajectory-based retrieval can recover a substantial portion of useful preference information without requiring explicit profile construction. The relatively small gap, despite the absence of privileged signals, highlights the effectiveness of our approach as a practical alternative for real-world deployment. Although the trajectory collector (Qwen-3-VL-32B-Thinking) shares a model
family with one evaluated backbone, Qwen-3-VL-32B-Thinking obtains the
lowest \ouragentname success rate (10.42\%), suggesting that any
model-family-specific stylistic advantage does not directly translate into
higher benchmark performance.

\paragraph{User-Centric and Oracle Agents.}
The User-Centric agent, which has access to the complete gold user profile, substantially improves performance across all models, reaching 68.33\% for Gemini-3-Pro. This demonstrates the upper bound achievable when full user information is available, though the agent must still identify which preferences are relevant to each task. The Oracle agent, which is given direct access to the single task-relevant preference, further increases performance to as high as 82.92\%. The large gap between these upper bounds and all other settings underscores that the primary challenge lies not only in acquiring user preferences, but in accurately selecting and applying the task-relevant subset at inference time. Further analysis is in \ref{sec:orac_limit}.

\subsubsection{Preference Alignment Score}

\begin{wraptable}{r}{0.45\linewidth}
\vspace{-25pt}
\centering
\centering
\caption{Preference alignment score across six models.}
\label{tab:prefer_align}
\begin{tabular}{lcc}
\toprule
Model & \ouragentname \\
\midrule
Gemini-3-Pro        & 27.71\% \\
Gemini-3-Flash      & 26.67\% \\
GPT-5.4             & 23.96\% \\
GPT-5-mini          & 29.79\% \\
Qwen-3.5-27B   & 22.50\% \\
Qwen-3-VL-32B   & 24.58\% \\
\bottomrule
\end{tabular}
\vspace{-15pt}
\label{tab:prefer_align}
\end{wraptable}

In addition to task success, we evaluate each model using \emph{Preference Alignment Score}. One of the most frequent error cases arises when the task requires understanding the user’s product filtering preference, but the agent instead infers a product sorting preference (e.g., sorting by price or name), leading it to sort the products instead of filtering them according to the user’s constraints.

Across all backbone models, preference alignment scores (22.50\%–29.79\%) are consistently higher than the corresponding task success rates under \ouragentname (10.42\%–15.62\%), revealing a systematic gap between \emph{preference understanding} and \emph{preference-grounded task execution}. While models often correctly infer the user’s intended preference, they frequently fail to operationalize it into successful interactions within the environment.
This gap is particularly pronounced for GPT-5-mini, as it achieves the highest preference alignment score (29.79\%) but only attains an 11.67\% task success rate, indicating that correctly inferred preferences do not reliably translate into successful task completion. Similarly, Qwen-3-VL-32B-Thinking exhibits a preference alignment score of 24.58\% but achieves only 10.42\% success, suggesting substantial execution failures despite partial preference understanding.

To assess the reliability of the Preference Alignment Score, we conduct a human--LLM agreement study on a subset of episodes. The two human annotators exhibit near-perfect agreement (Cohen's $\kappa = 0.93$). The LLM judge also demonstrates strong agreement with human judgments (Cohen's $\kappa = 0.80$ and $0.87$ against the two human annotators, respectively), supporting the use of LLM-based preference alignment as a reliable evaluation signal. Study details are provided in Appendix \ref{sec:c2}.

\section{Analysis}
\label{sec:analysis}

\subsection{Single vs. Compositional Preferences}

\begin{figure}[t]
    \centering
    
    \begin{minipage}[t]{0.48\linewidth}
        \centering
        \includegraphics[width=\linewidth]{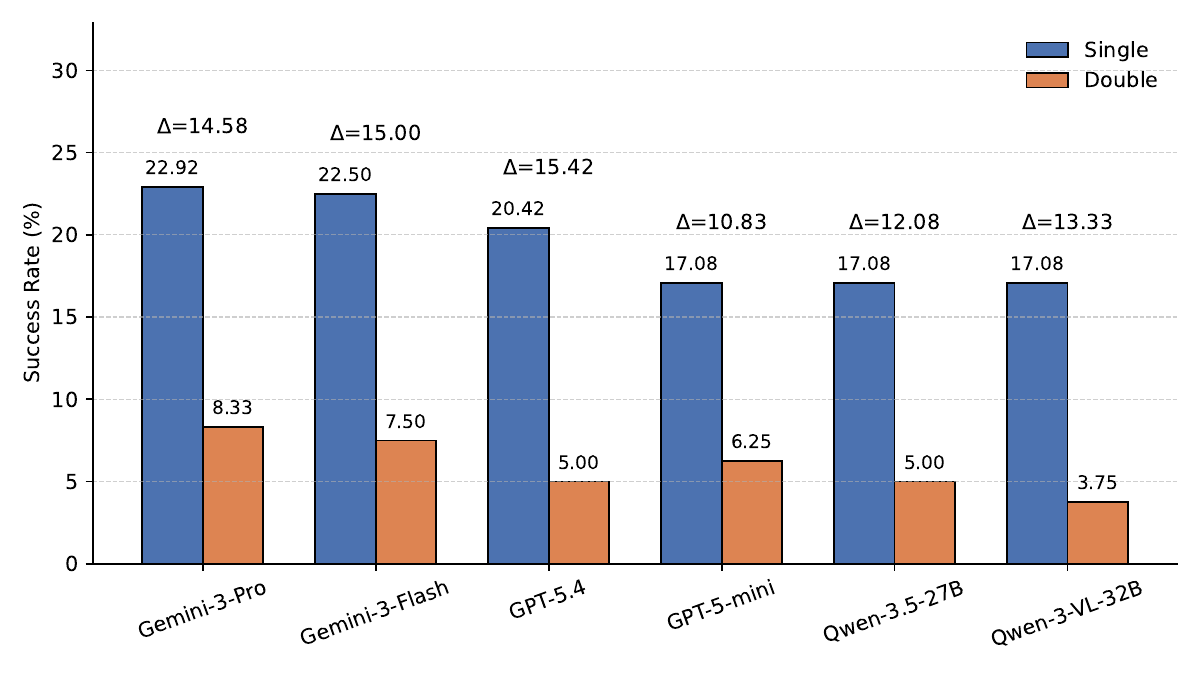}
        \captionof{figure}{Task success rate of each model under single and double preferences.}
        \label{fig:success}
    \end{minipage}
    \hfill
    \begin{minipage}[t]{0.48\linewidth}
        \centering
        \includegraphics[width=\linewidth]{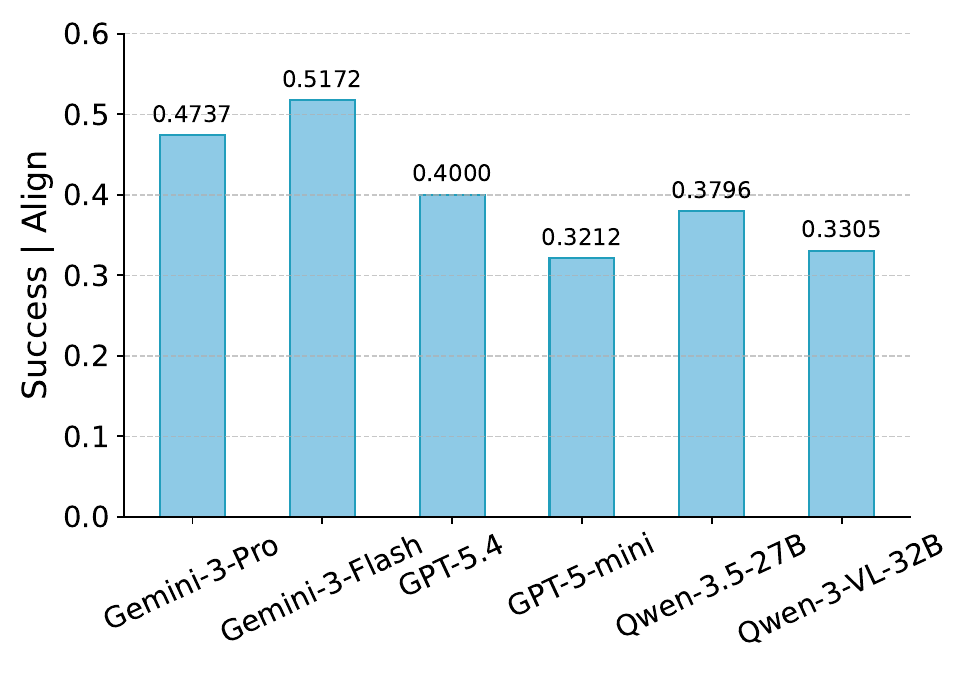}
        \captionof{figure}{Conditional success rate across models.}
        \label{fig:align}
    \end{minipage}
    
\end{figure}

Figure \ref{fig:success} shows the performance of \ouragentname across single-preference and double-preference tasks. Across all backbone models, we observe a consistent and substantial degradation when moving from single-preference to double-preference tasks. For instance, GPT-5.4 achieves 20.42\% success on single-preference tasks but drops sharply to 5.00\% on double-preference tasks, yielding a gap of 15.42 percentage points. Similar trends hold for other models: Gemini-3-Pro decreases from 22.92\% to 8.33\%, GPT-5-mini from 17.08\% to 6.25\%, and Qwen-3-VL-32B-Thinking from 17.08\% to 3.75\%. Performance remains challenging across domains: ~\ref{app:disaggregated} shows no domain--task-type cell exceeds 33\% success, indicating that the difficulty is not confined to a particular environment. The single-to-double preference degradation is also consistent across domains and backbone models. While the absolute performance varies across backbones and domains, with Gemini-3-Pro generally achieving higher success on Map and Shopping tasks, the same qualitative trend persists across models.

This pattern reveals two key insights. First, compositional preference understanding remains a major bottleneck for current web agents. While models demonstrate limited ability to infer and apply a single latent preference, they struggle significantly when required to jointly reason over multiple constraints. Second, the magnitude of the gap, ranging from 10.83 to 15.42 percentage points across models, suggests that this limitation is systematic rather than model-specific.

A representative example of successful compositional alignment shows the agent correctly integrating both a driving preference and a location profile. It explicitly reasons that \textit{the transport mode is set to Car (OSRM), which matches the user's preference}, while also grounding the task in \textit{the user's location, Lucas County, Ohio}. Crucially, this reasoning is translated into action: when automatic detection fails, the agent manually inputs the location, demonstrating that both preferences actively guide decision-making rather than being passively satisfied. Ultimately, these findings underscore compositional personalization as a critical and underexplored challenge.

\subsection{Retrieval Quality}
\label{sec4.2}
The \textsc{AdaptiveAgent} pipeline acquires a preference in three stages: retrieving relevant trajectories, extracting a preference, and executing the task under that preference. Section~\ref{sec4.3} analyzes the latter two stages; here, we measure retrieval directly. Retrieval uses a frozen CLIP text query and FAISS inner-product search, making it deterministic and identical across all six backbones.
We define retrieval recall at the preference-category level. Each user encodes twelve preferences, one per category and trajectory. Retrieval succeeds when the top-$K$ trajectories include the trajectory corresponding to the task's gold-preference category. The agent retrieves $K{=}2$ trajectories, so recall@2 is the operating point inherited by downstream stages.
Recall@2 is 71.88\% overall, 77.50\% on single-preference tasks, and 66.25\% on double-preference tasks; recall@1 is 59.17\%. In the full pipeline, retrieval recovers the correct trajectory 71.88\% of the time, but preference alignment is only 22.50--29.79\% (Table~\ref{tab:prefer_align}), and task success is 10.42--15.62\% (Table~\ref{tab:main_result}). Thus, most of the loss occurs after retrieval, in preference extraction and execution. This also explains why \textsc{AdaptiveAgent} approaches the Reconstructed Profile baseline despite its stricter setting: downstream stages discard much of the signal that retrieval has already recovered.

Retrieval quality varies substantially across preference categories (Table~\ref{tab:retrieval_quality}). Location, Phone, Diet, Brand, and Reddit Forum achieve 100\% recall@2, whereas Product Sorting and Wishlist vs.\ Cart achieve only 35.0\% and 5.0\%, respectively. The latter reflects a limitation of visual retrieval: sorting, filtering, and wishlist/cart actions can produce similar rendered states and differ mainly in the control action rather than the page appearance. Notably, Wishlist vs.\ Cart is retrieved only 5.0\% of the time, yet achieves 63.3\% alignment once the relevant trajectory is retrieved.

Separating retrieval from downstream stages explains the near-zero categories in Figure~\ref{fig:preference_alignment_comparison222}. Product Filtering is retrieved 75.0\% of the time but aligns only 3.3\%, indicating a downstream failure, whereas Wishlist vs.\ Cart is retrieved only 5.0\% but aligns 63.3\% once conditioned, indicating a retrieval failure. These contrasting cases show why retrieval must be measured separately: low task performance can arise because a preference is not retrieved or because it cannot be effectively used after retrieval, pointing to richer retrieval signals or stronger action grounding, respectively.

\subsection{Preference Alignment vs. Execution Gap}
\label{sec4.3}
\begin{wraptable}{r}{0.48\textwidth}
\vspace{-10pt}
\centering
\caption{Execution-level failure categories among unsuccessful episodes with correct preference alignment. AF: Action Failure, TF: Termination Failure, NL: Navigation Loop.}
\scriptsize
\setlength{\tabcolsep}{6pt}
\begin{tabular}{lrrrr}
\toprule
Model & AF & TF & NL & Other \\
\midrule
Gemini-3-Pro & 21.43\% & 44.29\% & 27.14\% & 7.14\% \\
Gemini-3-Flash & 35.29\% & 29.41\% & 30.88\% & 4.41\% \\
GPT-5.4 & 37.68\% & 34.78\% & 21.74\% & 5.80\% \\
GPT-5-mini & 46.75\% & 35.06\% & 14.29\% & 3.90\% \\
Qwen-3.5-27B & 38.81\% & 46.27\% & 10.45\% & 4.48\% \\
Qwen-3-VL-32B & 40.51\% & 39.24\% & 12.66\% & 7.59\% \\
\bottomrule
\end{tabular}
\label{tab:failure_breakdown}
\vspace{-10pt}
\end{wraptable}

To isolate the source of the performance gap, we examine task success conditioned on correct preference alignment. Even when the inferred preference matches the ground truth, success rates remain limited: 47.37\% for Gemini-3-Pro, 40.00\% for GPT-5.4, and 37.96\% for Qwen-3.5-27B, as shown in Figure~\ref{fig:align}. This reveals a consistent pattern: \textit{correctly inferring user preferences is necessary but not sufficient for successful task completion.} Even when preference alignment is correct, execution failures remain common, indicating that downstream interaction control constitutes a distinct bottleneck beyond preference inference.

To better understand these failures, we categorize unsuccessful aligned trajectories into four execution-level error types in Table~\ref{tab:failure_breakdown}: \textbf{Action Failure}, where the intended UI interaction fails; \textbf{Navigation Loop}, where the agent repeatedly revisits similar states without progress; \textbf{Termination Failure}, where the agent fails to stop appropriately; and \textbf{Other}, which covers execution failures that do not fit the above categories. The distribution of failure modes varies substantially across models. Termination Failure and Navigation Loop are prominent for several models, while Action Failure accounts for a substantial fraction of failures in others. For example, Termination Failure is the largest category for Gemini-3-Pro (44.29\%) and Qwen-3.5-27B (46.27\%), whereas Action Failure is most frequent for GPT-5-mini (46.75\%), GPT-5.4 (37.68\%), and Qwen-3-VL-32B (40.51\%). Gemini-3-Flash exhibits a more balanced distribution across Action Failure (35.29\%), Termination Failure (29.41\%), and Navigation Loop (30.88\%). These failures can occur even when the agent has correctly inferred the relevant user preference. For example, an agent may correctly infer that products should be sorted alphabetically, but fail to complete the task due to repeated navigation, unsuccessful UI interactions, or failure to terminate after obtaining the required information. In some cases, the agent explicitly recognizes the execution problem, e.g., \textit{``we were not able to get the first product (the user’s request) because the previous click attempts failed.''} The most notable failure from \textit{Other} category is environment instability, such as repeated page-load failures that prevent successful task execution. Overall, these findings suggest that downstream execution remains a major bottleneck even when the relevant user preference has been correctly inferred.

\section{Ablation}
\label{sec:ablation}

\subsection{Cross-User Trajectory Assignment}

\begin{wraptable}{r}{0.45\linewidth}
\vspace{-25pt}
\centering
\centering
\caption{Success rate under cross-user trajectory assignment.}
\label{tab:cuta}
\begin{tabular}{lcc}
\toprule
Model & \ouragentname\\
\midrule
Gemini-3-Pro & 7.50\% \\
GPT-5-mini & 7.71\% \\
Qwen-3.5-27B & 8.96\%\\
\bottomrule
\end{tabular}
\vspace{-15pt}
\label{tab:cuta}
\end{wraptable}

To test whether performance gains depend on correctly aligned user context, we evaluate the agent under a \emph{cross-user trajectory assignment} setting. Here, the agent is conditioned on trajectories $\{T_j\}$ generated from user $U_j$ but evaluated on tasks $X_i$ associated with a different user $U_i$. This preserves the trajectory distribution and task format while explicitly removing the correspondence between user intent and conditioning context.

To further stress-test this setting, we deliberately swap trajectories between users with the most contradictory preferences. As shown in Table~\ref{tab:cuta}, this misalignment consistently leads to a substantial drop in success rate across all models, reducing performance close to the no profile baseline in Table~\ref{tab:main_result}. If trajectory conditioning mainly captured task overlap or dataset artifacts, using mismatched trajectories should still provide some benefit. The degradation is observed across both Gemini-3-Pro and Qwen-3.5-27B,
with success dropping from 15.62\% to 7.50\% and from 11.04\% to
8.96\%, respectively. These results preserve the trajectory distribution
and task format while disrupting only the correspondence between user
preferences and historical trajectories, providing evidence that the gains
depend on correctly aligned user-specific signals rather than merely
retrievable task overlap or generic trajectory artifacts.

\section{Related Works}
\label{sec:related}
\paragraph{Benchmarks for Digital Agents}
Recent advances in Large Language Models have enabled agents to operate across diverse virtual environments, including web browsers~\citep{yao2022webshop, lu2024weblinx, qi2025webrltrainingllmweb}, mobile devices~\citep{rawlesandroidworld, ye2025mobile}, and operating systems~\citep{wang2025opencua, xie2025scalingcomputerusegroundinguser, lu2025videoagenttrekcomputerusepretraining}, supported by specialized tools and evaluation harnesses~\citep{chezelles2025browsergym, workarena2024}. In parallel, a growing body of work has explored general-purpose LLM-based agents capable of planning and tool use, such as \cite{yao2022react, schick2023toolformer, qin2023toolllm, wang2023voyager}, as well as embodied and interactive agents operating in simulated environments~\citep{shridhar2020alfred, mu2023embodiedgpt}.
Despite this progress, existing benchmarks ~\citep{zhou2023webarena, deng2023mind2web, xie2024osworld, osworld_verified, wang2025computeragentarena} focus on task success rates, with limited attention to user preferences.

\paragraph{Agent Personalization}
Research on personalized agents has evolved from explicit user profiles toward retrieval-augmented generation~\citep{salemi2024lamplargelanguagemodels, richardson2023integratingsummarizationretrievalenhanced} and memory-based approaches~\citep{li2025hello, su2026beyond} that incorporate long-term user context. Another research~\citep{li2025hello,wu2025aligning,zhao2025llms, kim2025cupid, zhang2025personaagent} works on user preference alignment through multi-turn dialogues from interactions. General User Models~\citep{shaikh2025creatinggeneralusermodels} present a general architecture that infers user knowledge and preferences from multimodal computer-use observations. In the web domain, benchmarks like PersonalWAB~\citep{cai2025large} and Persona2Web~\citep{kim2026persona2web} have begun to evaluate personalization capabilities. A key open challenge remains enabling agents to infer and apply task-relevant implicit preferences from historical trajectories at test time.

\label{sec:conclusion}
\section{Conclusion}

We introduce \ada, a benchmark for evaluating test-time personalization of web agents via implicit preference inference from historical user trajectories. To validate its difficulty, we propose \ouragentname, a simple framework that infers latent preferences from past interactions and integrates them into decision-making. Despite access to historical trajectories, \ouragentname achieves less than half the performance of an oracle with ground-truth preferences, identifying implicit preference inference as a key bottleneck in long-horizon web interactions. We further find that accurate preference inference alone is insufficient for successful execution, highlighting the need for robust downstream decision-making and action grounding.

\subsection*{AI use statement}

In this work, we used generative AI tools for polishing and improving the clarity and readability of the manuscript text, as well as for generating code used in our experiments. We have not used generative AI tools for developing research ideas, designing the methodology, analyzing or interpreting experimental results, or generating research claims. We reviewed all AI-assisted text for accuracy and consistency with the underlying work, and all AI-generated code was manually reviewed, tested, and verified for correctness. We take responsibility for the final content of this work, including all text, claims, code, and other research artifacts produced with the aid of generative AI.

\subsection*{Reproducibility statement}
We provide detailed descriptions of our proposed framework and experimental setup in the main paper, including trajectory collection, deployment task annotation and evaluation protocols. Additional implementation details, hyperparameters, and experimental results are provided in the appendix and supplementary materials. We will also provide the source code and necessary configuration files to facilitate reproduction of our experiments.

\bibliography{iclr2027_conference}
\bibliographystyle{iclr2027_conference}

\appendix
\appendix
\label{sec:ablation}
\FloatBarrier
\section{Limitation}

\ada has several limitations that define the scope of our conclusions. 

First, our evaluation is conducted in three controlled web environments from WebArena through BrowserGym. While these environments provide reproducible and interactive settings, they do not capture the full diversity of real-world websites, including dynamically changing interfaces, longer-term user interactions, or domains with richer personalization signals. Second, historical trajectories are constructed under explicitly specified persona traits rather than collected from naturally occurring long-term user behavior. This controlled construction enables systematic evaluation and precise ground-truth preferences, but may not capture the ambiguity, inconsistency, and sparsity of preferences expressed in real user histories. Third, our benchmark primarily models preferences as relatively stable signals that can be inferred from past behavior. It therefore does not evaluate settings in which preferences evolve over time or depend strongly on context, task, or situational constraints. Fourth, \ouragentname uses a specific screenshot-based retrieval design with CLIP embeddings and retrieves two trajectory prefixes. Finally, the trajectory collection model, Qwen-3-VL-32B-Thinking, belongs to the same model family as one evaluated backbone, introducing a potential source of stylistic bias in the collected histories. Although Qwen-3-VL-32B-Thinking achieves the lowest \ouragentname success rate among the evaluated backbones, ruling out such bias would require collecting trajectories with diverse models or, ideally, real users. Addressing these limitations with naturally occurring user histories, broader environments, dynamic preferences, and model-diverse trajectory collection is an important direction for future work.

\section{Benchmark}
\label{sec:benchmark}
\subsection{Sample Trajectories}
\subsubsection{Success Case}
We provide a sample successful trajectory that captures the user's preference. The ground-truth user preference in this example is a preference for filtering Health \& Household section in Health Care category.

\textbf{Task Instruction}: I prefer buying Health \& Households products from Health Care category. Can you tell me the price of the first product listed?
\begin{figure}[ht]
    \centering

    \begin{subfigure}{0.48\textwidth}
        \centering
        \includegraphics[width=\linewidth]{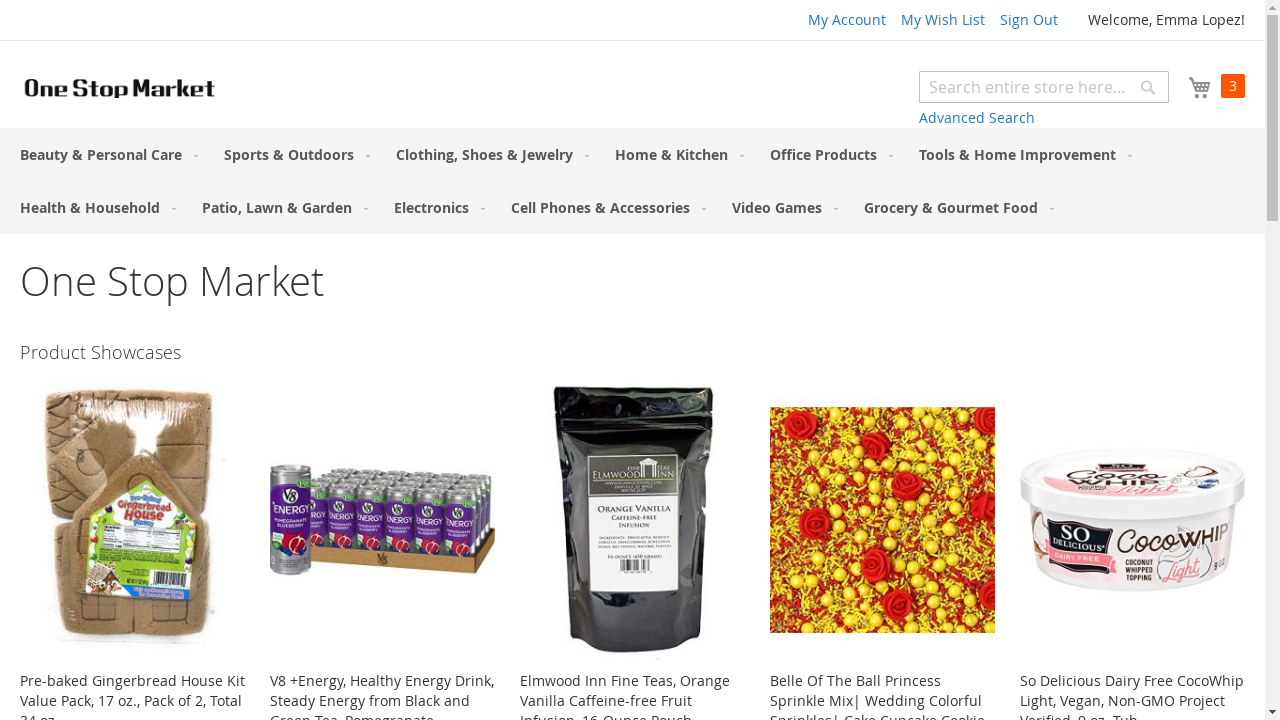}
        \caption{Step 1}
    \end{subfigure}
    \hfill
    \begin{subfigure}{0.48\textwidth}
        \centering
        \includegraphics[width=\linewidth]{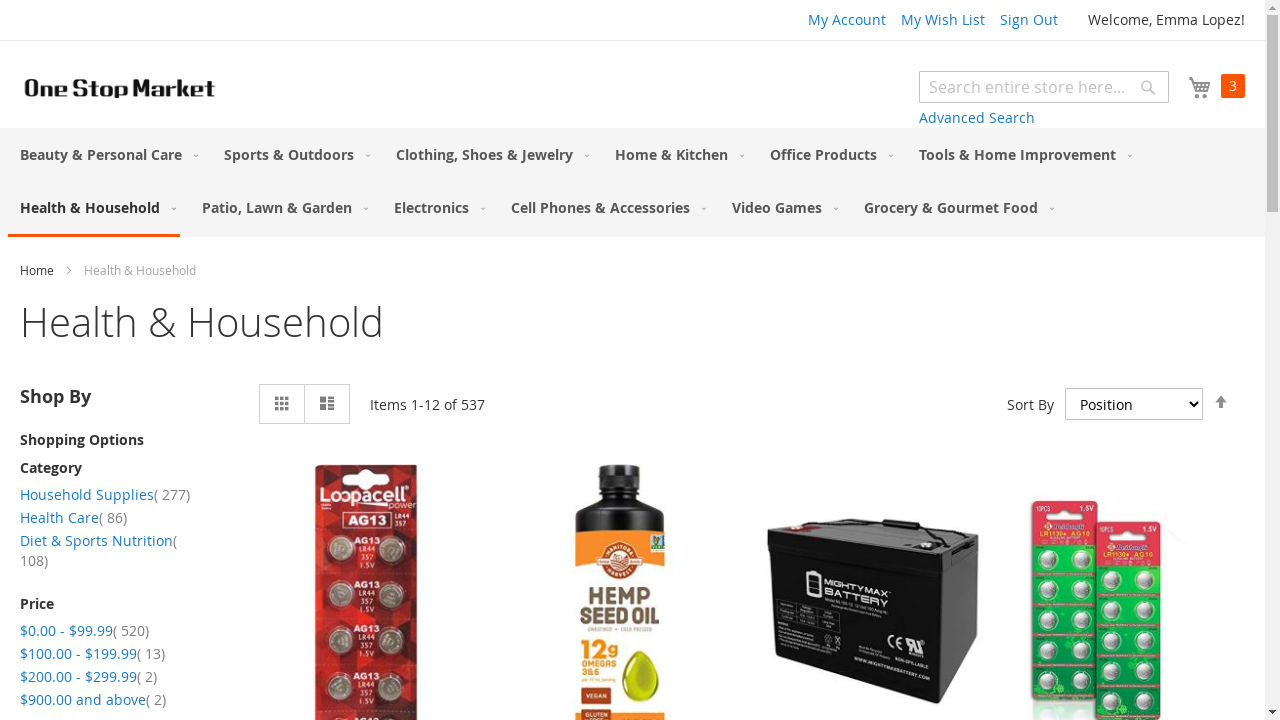}
        \caption{Step 2}
    \end{subfigure}

    \vspace{0.4cm}

    \begin{subfigure}{0.48\textwidth}
        \centering
        \includegraphics[width=\linewidth]{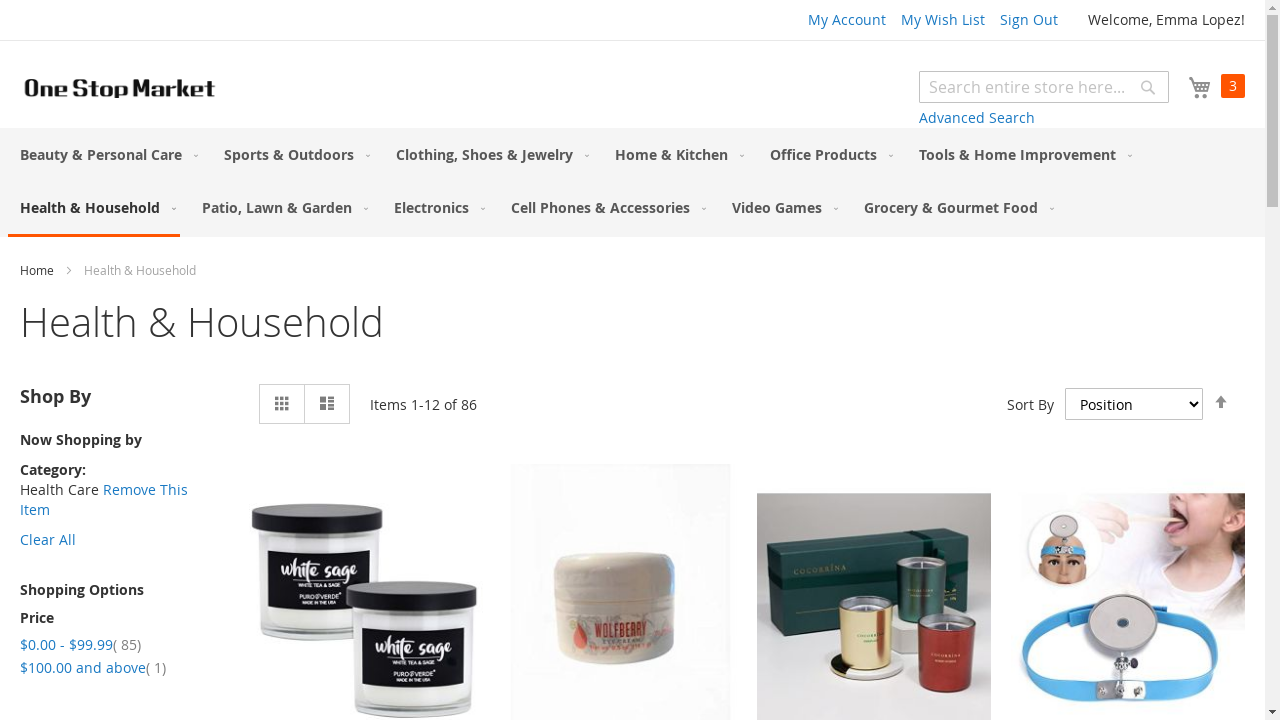}
        \caption{Step 3}
    \end{subfigure}
    \hfill
    \begin{subfigure}{0.48\textwidth}
        \centering
        \includegraphics[width=\linewidth]{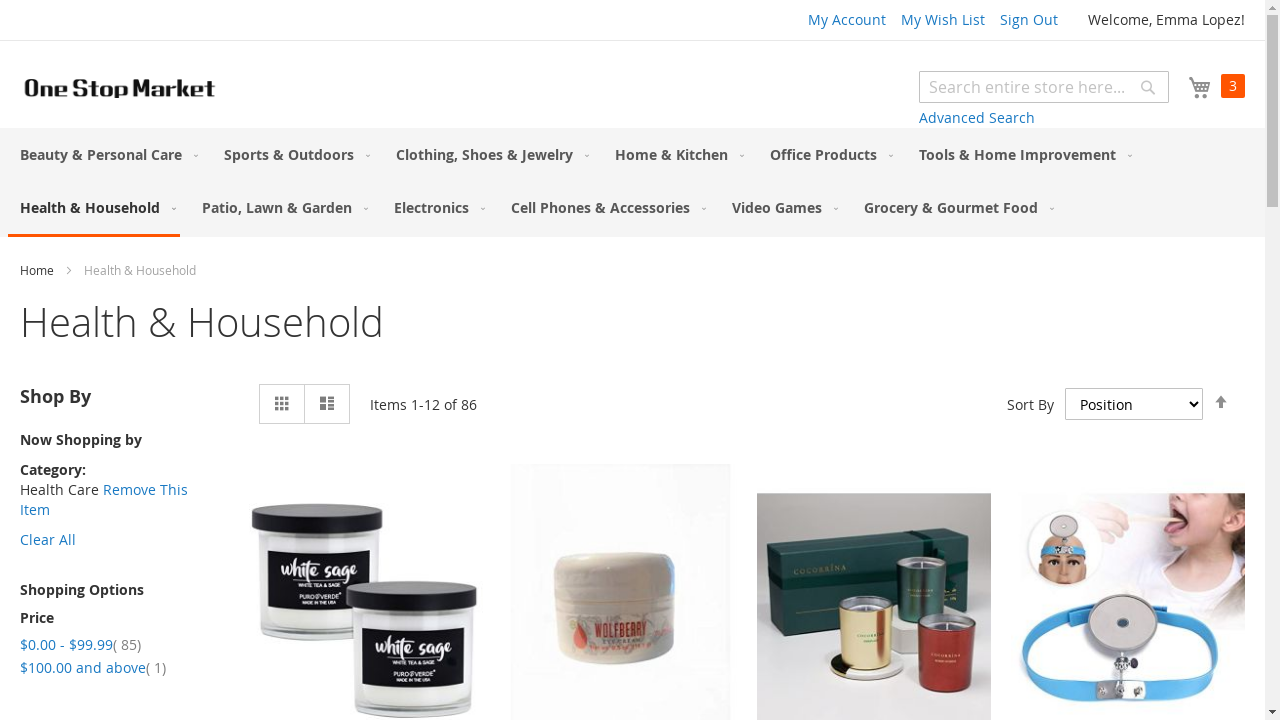}
        \caption{Step 4}
    \end{subfigure}

    \vspace{0.4cm}
    \caption{Sample successful trajectory shown as screenshot sequences.}
    \label{fig:success_traj}
\end{figure}
\FloatBarrier

\FloatBarrier
\subsubsection{Failed Case}
We also present a failed case in which the user preference is still correctly captured. The ground-truth user preference in this example is a preference for television Reddit forums.

\textbf{Task Instruction}: I only use the television forum. Who wrote the second most active submission?

\begin{figure}[!htb]
    \centering

    \begin{subfigure}{0.48\textwidth}
        \centering
        \includegraphics[width=\linewidth]{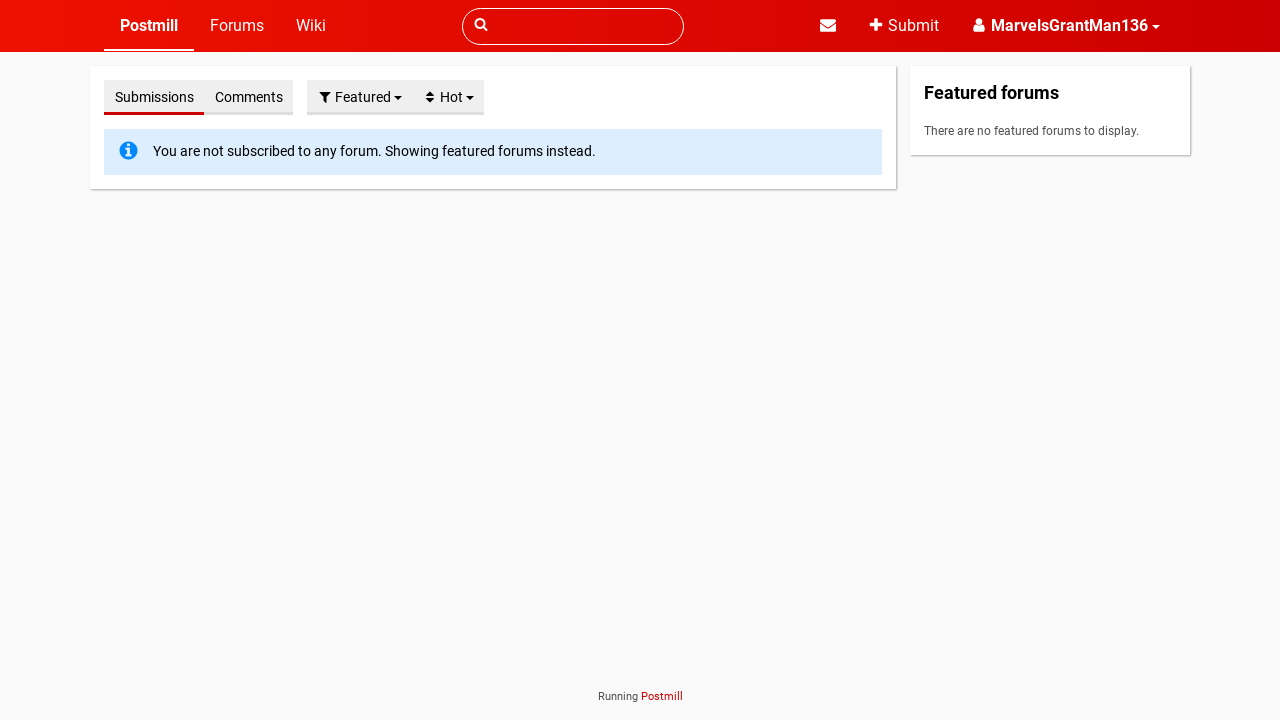}
        \caption{Step 1}
    \end{subfigure}
    \hfill
    \begin{subfigure}{0.48\textwidth}
        \centering
        \includegraphics[width=\linewidth]{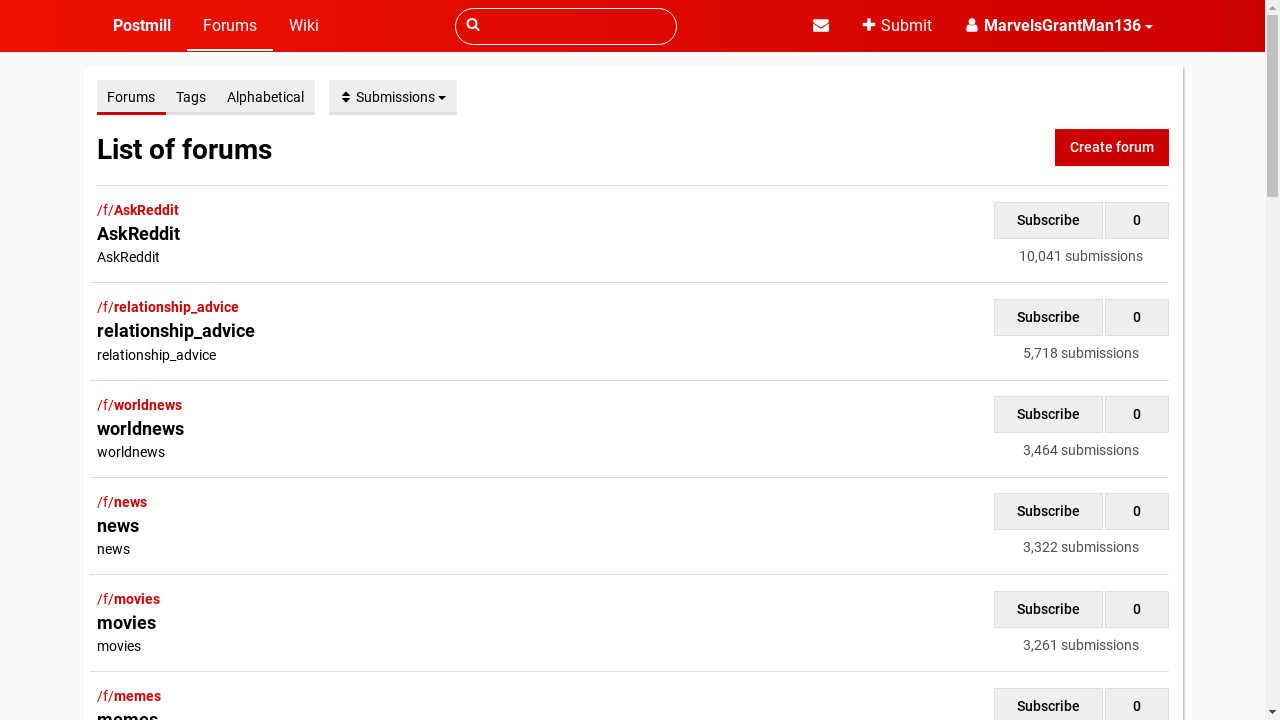}
        \caption{Step 2}
    \end{subfigure}

    \vspace{0.4cm}

    \begin{subfigure}{0.48\textwidth}
        \centering
        \includegraphics[width=\linewidth]{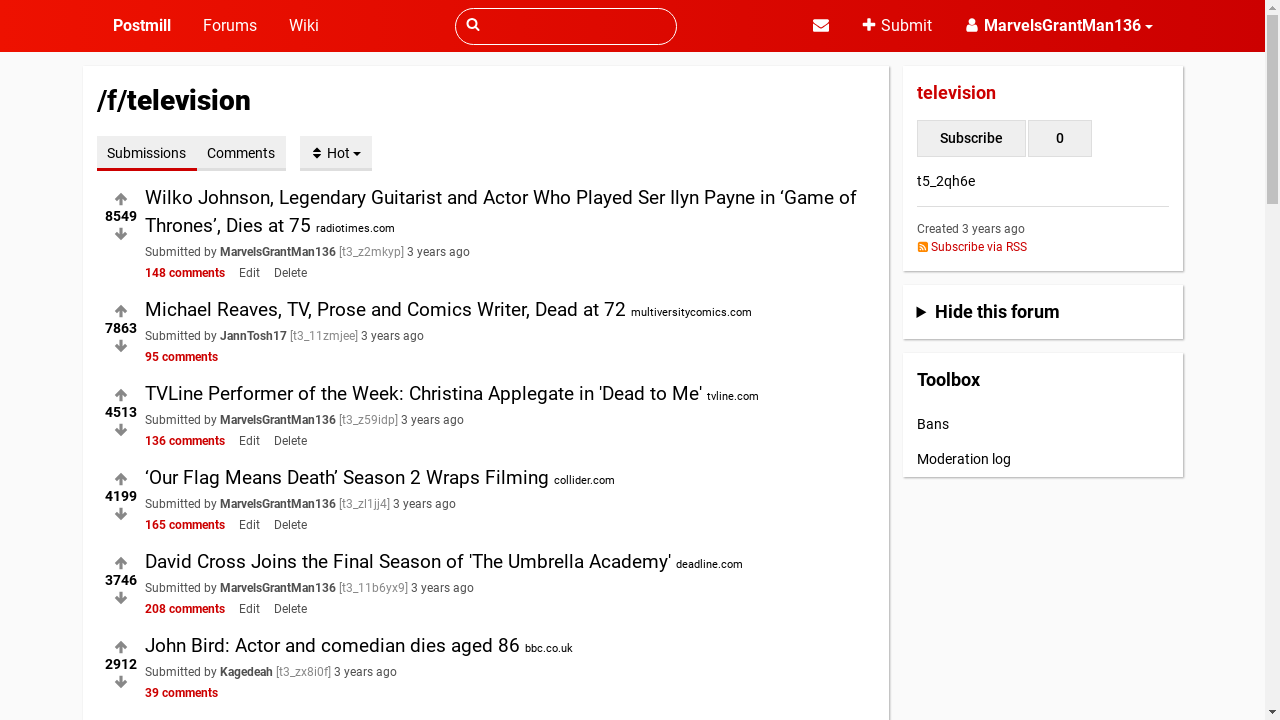}
        \caption{Step 3}
    \end{subfigure}
    \hfill
    \begin{subfigure}{0.48\textwidth}
        \centering
        \includegraphics[width=\linewidth]{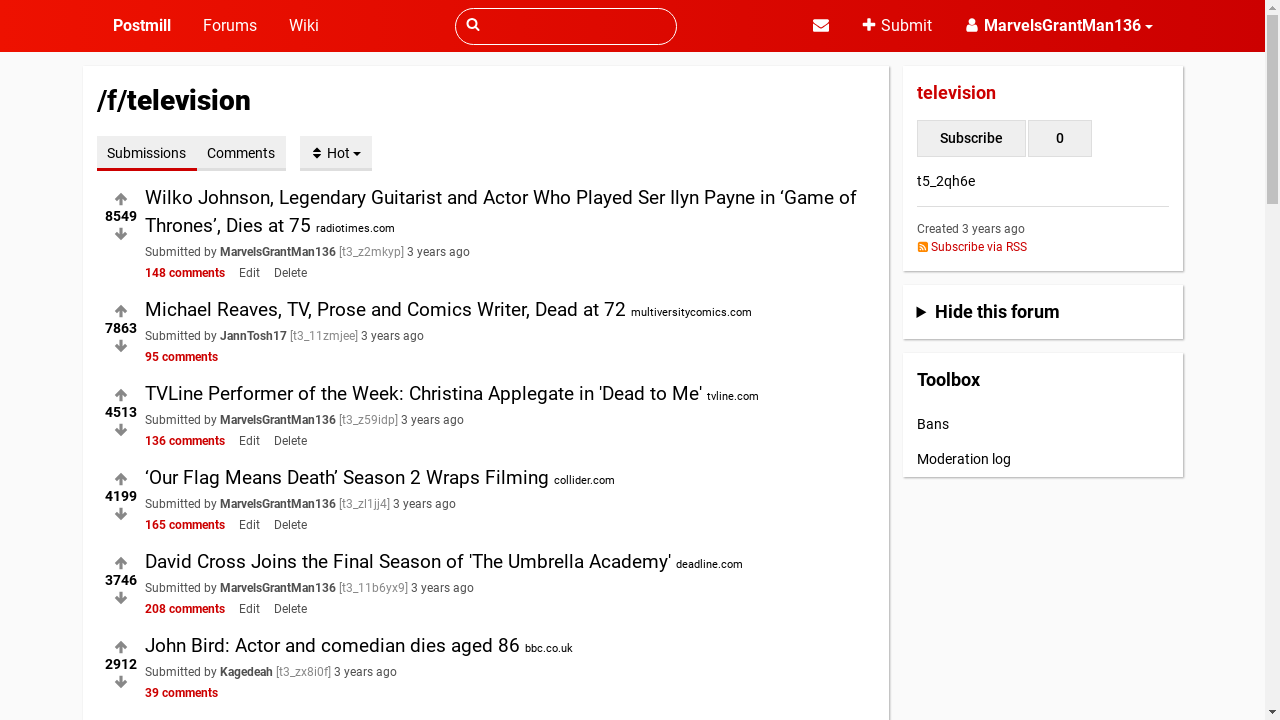}
        \caption{Step 4}
    \end{subfigure}

    \vspace{0.4cm}
    \caption{Sample failed trajectory shown as screenshot sequences.}
    \label{fig:fail_traj}
\end{figure}
\FloatBarrier

\subsection{Sample User Persona}
\label{sec:b1}

\begin{table}[htbp]
\caption{Sample profile.}
\label{tab:user01_profile}
\centering
\begin{tabular}{p{5cm} p{8cm}}
\toprule
Preference Type & Value \\
\midrule
Product Sorting Preference & Ascending Product Name Order \\
\midrule
Product Filtering Preference & Video Game filtering to XBox One Category \\
\midrule
Wishlist vs. Cart Preference & Wishlist \\
\midrule
Buying Preference & Lightweight product \\
\midrule
Regional Product Preference & Korean \\
\midrule
Brand Preference & Hebhac Herbs \\
\midrule
Transportation Preference & Bicycle \\
\midrule
Reddit Forum Preference & Movies \\
\midrule
Reddit Forum Sorting Preference & Active \\
\midrule
Diet Preference & Vegan \\
\midrule
My Phone Model & Galaxy Z Flip 3 \\
\midrule
My Location & Holmes County, Mississippi \\
\bottomrule
\end{tabular}
\end{table}

\begin{table}[htbp]
\caption{Sample profile.}
\label{tab:user03_profile}
\centering
\begin{tabular}{p{5cm} p{8cm}}
\toprule
Preference Type & Value \\
\midrule
Product Sorting Preference & Descending Product Name Order \\
\midrule
Product Filtering Preference & Office products filtering to Office \& School supplies category \\
\midrule
Wishlist vs. Cart Preference & Wishlist \\
\midrule
Buying Preference & Gaming product \\
\midrule
Regional Product Preference & France \\
\midrule
Brand Preference & LG \\
\midrule
Transportation Preference & Walk \\
\midrule
Reddit Forum Preference & Books \\
\midrule
Reddit Forum Sorting Preference & Active \\
\midrule
Diet Preference & non-GMO \\
\midrule
My Phone Model & Galaxy Note 9 \\
\midrule
My Location & Lawrence County, Indiana \\
\bottomrule
\end{tabular}
\end{table}

\begin{table}[htbp]
\caption{Sample profile.}
\label{tab:user04_profile}
\centering
\begin{tabular}{p{5cm} p{8cm}}
\toprule
Preference Type & Value \\
\midrule
Product Sorting Preference & Ascending Price Order \\
\midrule
Product Filtering Preference & Health \& Households filtering to Health Care category \\
\midrule
Wishlist vs. Cart Preference & Cart \\
\midrule
Buying Preference & Eco-friendly \\
\midrule
Regional Product Preference & USA \\
\midrule
Brand Preference & Estee Lauder \\
\midrule
Transportation Preference & Bicycle \\
\midrule
Reddit Forum Preference & Wallstreetbets \\
\midrule
Reddit Forum Sorting Preference & New \\
\midrule
Diet Preference & High-protein \\
\midrule
My Phone Model & iPhone 13Pro Max \\
\midrule
My Location & Franklin County, Vermont \\
\bottomrule
\end{tabular}
\end{table}

\FloatBarrier
\subsection{Sample Trajectory collection and Deployment Tasks}
\label{sec:b2}
\begin{table}[htbp]
\caption{Sample single preference trajectory collection tasks from each domain.}
\label{tab:sample_traj_tasks}
\centering
\begin{tabular}{p{1.5cm} p{6cm} p{5.5cm}}
\toprule
\textbf{Domain} & \textbf{Task Instruction} & \textbf{Gold User Profile} \\
\midrule
Shopping &
I usually sort products by ascending price. Can you tell me the SKU code of the first product in Electronics category? &
Sorting products in ascending price order \\
\midrule
Shopping &
I only purchase Sephora products. Can you add the powder brush to cart? &
Prefers Sephora brand \\
\midrule
Map &
I live in Holmes County, Mississippi. How long does it take to get from here to lexington, mississippi by bicycle? &
Lives in Holmes County, Mississippi\\
\midrule
Map &
I'd rather walk. Could you check how long it takes to get from British Museum to Tate Modern? & Prefers walking \\
\midrule
Forum &
I only browse the Movies forum. Can you tell me the title of the first submission listed in the forum? &
Prefers Movies reddit forum \\
\midrule
Forum &
I only use the Books forum. How many years ago was the first submission on the list posted? &
Prefers Books forum \\
\bottomrule
\end{tabular}
\end{table}

\begin{table}[htbp]
\caption{Sample single preference deployment tasks from each domain.}
\label{tab:sample_dep_tasks}
\centering
\begin{tabular}{p{1.5cm} p{6cm} p{5.5cm}}
\toprule
\textbf{Domain} & \textbf{Task Instruction} & \textbf{Gold User Profile} \\
\midrule
Shopping &
Could you save the Body Glide Original Anti-Chafe Balm for me? &
Prefers to save products in shopping cart \\
\midrule
Shopping &
Could you search for waterproof phonecase for my phone, then tell me its manufacturer? &
Uses iPhone12 Pro \\
\midrule
Map &
Can you search how long does it take to get from Cornell Tech to the Empire State Building? &
Prefers driving \\
\midrule
Map &
How long does it take to get from here to Prairie View Elementary School, Oregon by walk? &
Lives in Dane County, Wisconsin \\
\midrule
Forum &
How many upvotes does the most active submission have? &
Prefers Space reddit forum \\
\midrule
Forum &
Can you tell me the number of comments for the first submission listed in the forum? &
Prefers Funny forum \\
\bottomrule
\end{tabular}
\end{table}

\begin{table}[htbp]
\caption{Sample double preference deployment tasks from each domain.}
\label{tab:sample_dep_tasks}
\centering
\begin{tabular}{p{1.5cm} p{6cm} p{5.5cm}}
\toprule
\textbf{Domain} & \textbf{Task Instruction} & \textbf{Gold User Profiles} \\
\midrule
Shopping &
Can you tell me the manufacturer of the first video game product in the list? &
1. Sorting products in ascending product name order

2. Filtering video game category into Xbox One subcategory \\
\midrule
Shopping &
Can you tell me the SKU code of the first electronics product in the list? &
1. Sorting products in ascending price order

2. Filtering electronics category into video projectors subcategory\\
\midrule
Map &
How long does it take for me to go from here to Murray Forest Park, Bedford? &
1. Prefers walking 

2. Lives in Lawrence County, Indiana\\
\midrule
Map &
How long does it take for me to go from here to Willis Day Industrial Park, Wood County? &
1. Prefers driving

2. Lives in Lucas County, Ohio \\
\midrule
Forum &
Can you tell me the title of the first forum submission in the list? &
1. Prefers Wallstreetbets reddit forum 

2. Prefers to sort reddit submissions in "New" Order\\
\midrule
Forum &
Can you tell me the name of the author of the third forum submission in the list? &
1. Prefers mildlyinteresting forum 

2. Sorting reddit submissions in "Active" Order\\
\bottomrule
\end{tabular}
\end{table}

\FloatBarrier
\subsection{Sample Reconstructed Profile}

\label{sec:b3}

\begin{figure*}[!htbp]
  \centering
\begin{tcolorbox}[colback=gray!10, colframe=black, title={Reconstructed profile for user01}]

I prefer vegan options and trusted brands (like Hebhac Herbs), and I often require products to be made in Korea. I usually research carefully by using search and name-ascending sorts, verifying SKUs, stock indicators, and origin details, then saving suitable items to my wishlist before buying. I tend to choose bicycling when looking up routes and rely on practical on-page cues (like breadcrumbs and Add to Cart buttons) to confirm context and availability.

\end{tcolorbox}
\caption{Reconstructed profile for user01.}
\end{figure*}

\begin{figure*}[!htbp]
  \centering
  
\begin{tcolorbox}[colback=gray!10, colframe=black, title={Reconstructed profile for user02}]

I prefer driving when getting directions and gravitate toward user-friendly products. I tend to be brand- and origin-conscious in shopping, buying only Sephora items in beauty and insisting on products made in Japan. I usually shop carefully with budget and details in mind—sorting by lowest price, checking stock status and SKUs, and selecting gluten-free options when relevant.

\end{tcolorbox}
\caption{Reconstructed profile for user02.}
\end{figure*}

\begin{figure*}[!htbp]
  \centering
\begin{tcolorbox}[colback=gray!10, colframe=black, title={Reconstructed profile for user03}]

I tend to search directly, sort and filter results, and focus on precise details like SKUs, counts, and timestamps before acting. I prefer trusted brands and clear quality markers (OEM accessories, non-GMO foods, and products made in France), and I often save items to a wishlist to compare them later. I usually shop for gaming gear and electronics alongside office supplies and gourmet items, and I often check walking times rather than driving when planning routes.

\end{tcolorbox}
\caption{Reconstructed profile for user03.}
\end{figure*}

\FloatBarrier
\subsection{User trajectory data statistics}
\label{subsec:traj}
We provide statistics for the user trajectory in table \ref{tab:4x2_example}, \ref{tab:action-distribution}, \ref{tab:trajectory-length}.
\begin{table}[htbp]
\caption{User trajectory statistics.}
\label{tab:4x2_example}
\centering
\begin{tabular}{cc}
\toprule
\textbf{Metric} & \textbf{Value} \\
\midrule
Trajectory Number & 120 \\
\midrule
Total Number of Steps & 1174 \\
\midrule
Average Total Steps per User & 117.40 \\
\midrule
Average Number of Steps across Trajectories & 9.78 \\
\bottomrule
\end{tabular}
\end{table}

\begin{table}[!htb]
\centering
\begin{minipage}[t]{0.55\textwidth}
\centering
\caption{Action types across the 120 collected trajectories.}
\label{tab:action-distribution}
\begin{tabular}{lrr}
\toprule
Action Type & Count & Fraction \\
\midrule
Click & 700 & 66.4\% \\
Fill & 137 & 13.0\% \\
No-op & 78 & 7.4\% \\
Scroll & 72 & 6.8\% \\
Select option & 36 & 3.4\% \\
Focus & 23 & 2.2\% \\
Goto / Go back / Type & 8 & 0.7\% \\
\bottomrule
\end{tabular}
\end{minipage}%
\hfill
\begin{minipage}[t]{0.4\textwidth}
\centering
\caption{Trajectory length distribution.}
\label{tab:trajectory-length}
\begin{tabular}{lr}
\toprule
Length & Count \\
\midrule
3--5 & 43 \\
6--9 & 46 \\
10--15 & 13 \\
16--29 & 5 \\
31 (max) & 13 \\
\bottomrule
\end{tabular}
\end{minipage}
\end{table}

\subsection{Task Distribution}
\label{tab:task_dist}
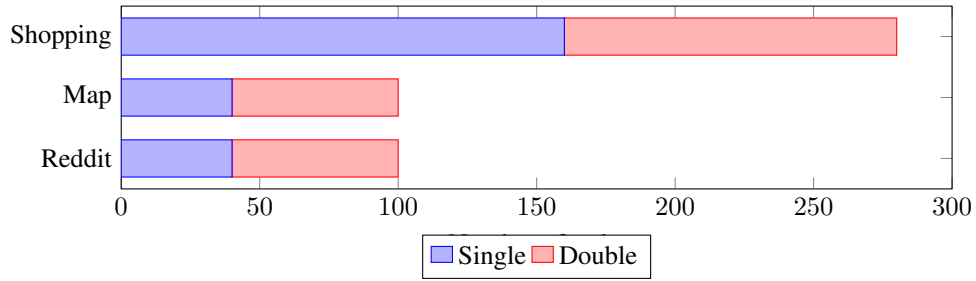
\begin{figure}[!htb]
    \centering
    \begin{tikzpicture}
        \begin{axis}[
            xbar stacked,
            bar width=14pt,
            width=0.9\linewidth,
            height=4.0cm,
            xmin=0,
            xmax=300,
            xlabel={Number of tasks},
            symbolic y coords={Reddit, Map, Shopping},
            ytick=data,
            xtick={0,50,100,150,200,250,300},
            enlarge y limits=0.25,
            legend style={
                at={(0.5,-0.25)},
                anchor=north,
                legend columns=2
            },
        ]

        \addplot coordinates {
            (40,Reddit)
            (40,Map)
            (160,Shopping)
        };

        \addplot coordinates {
            (60,Reddit)
            (60,Map)
            (120,Shopping)
        };

        \legend{Single, Double}
    \end{axis}
    \end{tikzpicture}
    \caption{Distribution of the 480 deployment tasks across web domains
    and preference composition.}
    \label{fig:task-distribution}
\end{figure}

\FloatBarrier
\section{Experiment details \& Additional results}

\subsection{Compute Resources}
All experiments conducted in this study, including those for the deployment phase and the \ouragentname framework, were executed using NVIDIA A100 (80GB) GPUs. The substantial memory capacity of the A100 was essential for handling the large-scale multimodal inputs required for web navigation and trajectory processing. On average, a complete experimental run for a single \ouragentname configuration across the benchmark tasks required approximately 10 to 12 hours of execution time.

\subsection{Human--LLM Agreement Analysis}
\label{sec:c2}
\paragraph{Human--LLM Agreement Analysis.}
To validate the reliability of the Preference Alignment Score, we conduct a human annotation study in which two annotators independently label whether an agent's interaction trajectory is behaviorally consistent with a specified user preference (binary labels: aligned or misaligned). Annotators are provided with the full interaction trajectory and preference description, and are blind to model identity and LLM judge outputs. We annotate 90 episodes sampled across three backbone models (Gemini-3-Pro, GPT-5.4, and Qwen-3.5-27B; 30 per model). Inter-annotator agreement is measured using Cohen's $\kappa$, yielding near-perfect human--human agreement ($\kappa = 0.93$). Agreement between the LLM-based judge and the two human annotators is also strong ($\kappa = 0.80$ and $0.87$, respectively), supporting the reliability of LLM-based preference alignment as an evaluation signal.

No monetary compensation was provided because the annotators were the authors.

\subsection{Limitations of User-Centric and Oracle Agents}
\label{sec:orac_limit}
Both User-Centric and Oracle Agents reveal complementary limitations in personalized web interaction. User-Centric Agents primarily fail at \emph{preference identification}: although they have access to the full ground-truth persona, they often struggle to determine which preference is relevant to the current task. Failures commonly arise from either over-applying multiple preferences or selecting irrelevant ones. For example, an agent may correctly recognize a preference for European cheese when searching for blue cheese, but additionally apply an unrelated product-sorting preference that alters the ranking and leads to an incorrect result. Similarly, when searching for a USB mouse, the agent may ignore the task-relevant gaming preference and instead prioritize a generic filtering heuristic. These cases show that richer personas alone do not guarantee better personalization; additional preferences can instead introduce more confounding signals. The core challenge is therefore not preference availability, but accurate \emph{preference discrimination and selective application}.

In contrast, Oracle Agents remove preference ambiguity by explicitly providing the relevant preference, yet still exhibit substantial execution-level failures, closely mirroring the issues discussed in the \emph{Preference vs.\ Execution Gap} analysis. Their errors stem primarily from weak UI grounding and brittle interaction strategies. Agents frequently rely on heuristic assumptions about interface structure (e.g., mapping “favorite” to Wish List or My Orders) without validating against the actual UI state, leading to repetitive navigation and looping behavior. They also tend to repeat failed interactions, assume nonexistent state changes, or continue re-checking pages even after the correct information has been found, reflecting poor adaptation and termination control.

\subsection{Preference Inference Difficulty Across Categories}

\begin{figure}[!htb]
    \centering
    \includegraphics[width=0.7\linewidth]{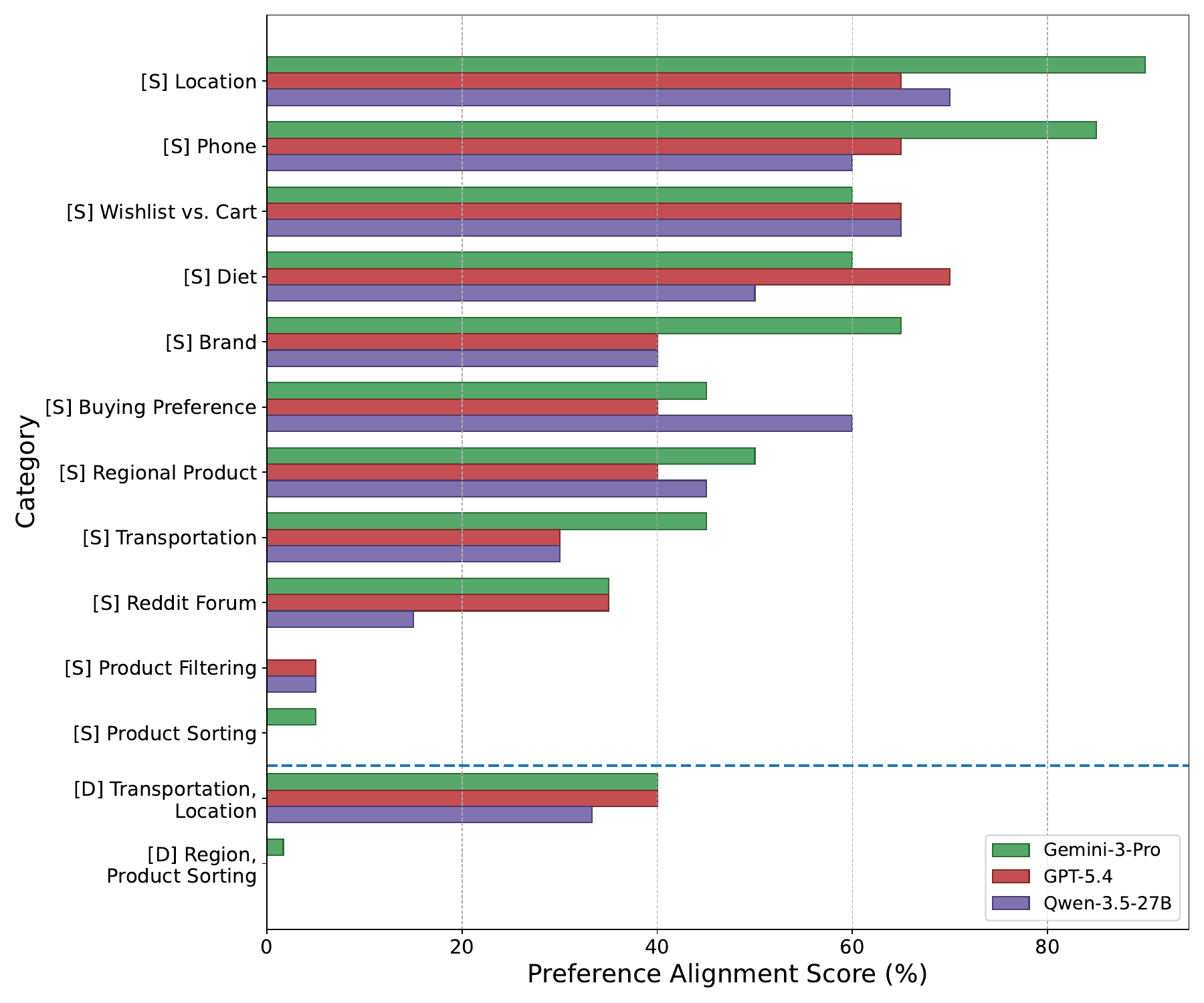}
    \caption{
Preference inference performance across categories. Explicit attribute-based preferences (e.g., \textit{Location}, \textit{Phone}) achieve consistently high success rates, while operation-oriented preferences (e.g., \textit{Product Sorting}, \textit{Product Filtering}) remain near zero across all models. Categories with 0\% alignment score across all models are omitted for clarity. [S]: Single, [D]: Double; separated by the dotted borderline
}
    \label{fig:preference_alignment_comparison222}
\end{figure}

Figure ~\ref{fig:preference_alignment_comparison222} shows preference inference performance across categories. 

Categories such as Location and Phone achieve the highest alignment score across all models. Gemini-3-Pro reaches 90\% on Location and 85\% on Phone, while GPT-5.4 and Qwen-3.5-27B also perform strongly (roughly 60–70\%). These preferences are grounded in clearly identifiable attributes, making them easier to extract and operationalize.

In contrast, Product Sorting, Product Filtering, and Reddit Forum Sorting are consistently the most challenging categories, with near-zero alignment score across all models. For example, Gemini-3-Pro achieves only 5\% on Product Sorting and 0\% on Product Filtering, while the remaining models almost entirely fail. These tasks require interpreting procedural intent—such as distinguishing sorting from filtering operations—which agents frequently confuse. This indicates a fundamental limitation in capturing action-level semantics rather than static user attributes.

The difficulty gap becomes even more pronounced in the double-preference setting. While relatively simple combinations such as Transportation + Location retain moderate performance (33–40\%), combinations involving structurally difficult preferences (e.g., Product Sorting + Product Filtering or Reddit Forum + Reddit Forum Sorting) collapse to near-zero across all models. This suggests that compositional preference inference is bottlenecked by the hardest constituent preference, highlighting the difficulty of jointly reasoning over heterogeneous user constraints.

\FloatBarrier
\subsection{Disaggregated Results}
\label{app:disaggregated}

\begin{table*}[!htb]
\centering
\caption{Task success rates (\%) by domain and preference composition.}
\label{tab:domain-task}
\begin{tabular}{llrrr}
\toprule
Domain & Model & Single & Double & Overall \\
\midrule
Shopping & Gemini-3-Pro & 25.0 & 7.5 & 17.5 \\
         & GPT-5.4 & 23.8 & 8.3 & 17.1 \\
         & Qwen-3.5-27B & 20.0 & 7.5 & 14.6 \\
\midrule
Map & Gemini-3-Pro & 32.5 & 18.3 & 24.0 \\
    & GPT-5.4 & 20.0 & 3.3 & 10.0 \\
    & Qwen-3.5-27B & 17.5 & 5.0 & 10.0 \\
\midrule
Reddit & Gemini-3-Pro & 5.0 & 0.0 & 2.0 \\
       & GPT-5.4 & 7.5 & 0.0 & 3.0 \\
       & Qwen-3.5-27B & 5.0 & 0.0 & 2.0 \\
\bottomrule
\end{tabular}
\end{table*}

\subsection{Retrieval Quality}

\begin{table}[t]
\centering
\caption{Retrieval recall@2 versus preference alignment by category (single-preference tasks). Retrieval recall is deterministic and backbone-independent; alignment is the mean over the three backbones of Figure~\ref{fig:preference_alignment_comparison222}. The two signals diverge by category: Wishlist vs.\ Cart fails at retrieval (5.0\%) yet aligns well once conditioned (63.3\%), whereas Product Filtering is retrieved well (75.0\%) yet almost never aligns (3.3\%).}
\label{tab:retrieval_quality}
\begin{tabular}{lcc}
\toprule
Category & Retrieval recall@2 & Alignment (mean) \\
\midrule
Location & 100.0\% & 75.0\% \\
Phone & 100.0\% & 70.0\% \\
Diet & 100.0\% & 60.0\% \\
Brand & 100.0\% & 48.3\% \\
Reddit Forum & 100.0\% & 28.3\% \\
Transportation & 95.0\% & 35.0\% \\
Reddit Sorting & 90.0\% & 0.0\% \\
Product Filtering & 75.0\% & 3.3\% \\
Regional & 80.0\% & 45.0\% \\
Buying & 50.0\% & 48.3\% \\
Product Sorting & 35.0\% & 1.7\% \\
Wishlist vs.\ Cart & 5.0\% & 63.3\% \\
\midrule
Overall recall@2 (single) & 77.5\% & \\
Overall recall@2 (all 480) & 71.9\% & \\
Recall@1 (single) & 59.2\% & \\
\bottomrule
\end{tabular}
\end{table}

\FloatBarrier
\subsection{Hyperparameters}
We use the same hyperparameter settings from BrowserGym ~\citep{chezelles2025browsergym}, with max\_prompt\_tokens set to 40K and html\_type set to pruned\_html. For all models, we use the temperature of 1.0.

\begin{table}[htbp]
\caption{Browsergym hyperparameters.}
\label{tab:hyperparams}
\centering
\begin{tabular}{p{12cm} p{2cm}} 
\toprule
\textbf{Hyperparameters} & \textbf{Value} \\
\midrule
use\_ax\_tree, use\_focused\_element, use\_error\_logs, use\_history, use\_action\_history, extract\_visible\_tag, use\_thinking, use\_concrete\_example, use\_abstract\_example, use\_hints, be\_cautious & True \\
\midrule
use\_html, use\_past\_error\_logs, use\_think\_history, use\_diff, use\_screenshot, use\_som, extract\_clickable\_tag, extract\_coords, filter\_visible\_elements\_only, use\_plan, use\_criticise, use\_memory, enable\_chat & False \\
\midrule
Maximum Number of Steps & 30 \\
\bottomrule
\end{tabular}
\end{table}
\FloatBarrier

\subsection{\ouragentname implementation details}
\label{sec:c4}

We provide additional implementation details of \ouragentname, an adaptive agent framework that separates preference inference from task execution via user-specific retrieval over past interaction trajectories. The framework maintains a database of trajectories for each user, where each trajectory is a sequence of timestamped screenshots ${s_0, s_1, \dots, s_T}$. At initialization, all screenshots are embedded using a pretrained CLIP ViT-B/32 model, and the resulting feature vectors are $\ell_2$-normalized and cached. For each user, we construct a FAISS inner-product index over these embeddings to enable efficient similarity search. 

At inference time, a task instruction is encoded into the same embedding space using the CLIP text encoder. We then perform sequential retrieval: at each step, we identify the single most similar screenshot in the remaining pool, recover its associated trajectory, and record the corresponding step index $k$. The selected trajectory is subsequently removed from the pool to encourage diversity, and the process is repeated for a fixed number of iterations. For each selected trajectory, we reconstruct a prefix consisting of all screenshots from $s_0$ to $s_k$, and aggregate prefixes from multiple trajectories into a unified set of images. These images are encoded in base64 format and provided as input to a multimodal preference extractor instantiated with GPT-5-nano. The model is prompted to infer the user’s latent preferences strictly based on observable interaction patterns, producing a concise first-person natural language summary (e.g., ``I prefer...''). The inferred preference is injected into the agent prompt as an auxiliary context block (\texttt{<image\_summary>}), which is concatenated with the current task instruction, observation history, and action space specification. The downstream web agent then conditions on both the task and inferred preference to generate actions. All embedding computations and FAISS indices are cached on disk to avoid recomputation, enabling scalable and efficient deployment across multiple users.

\subsection{Prompts}

\begin{figure*}[!htbp]
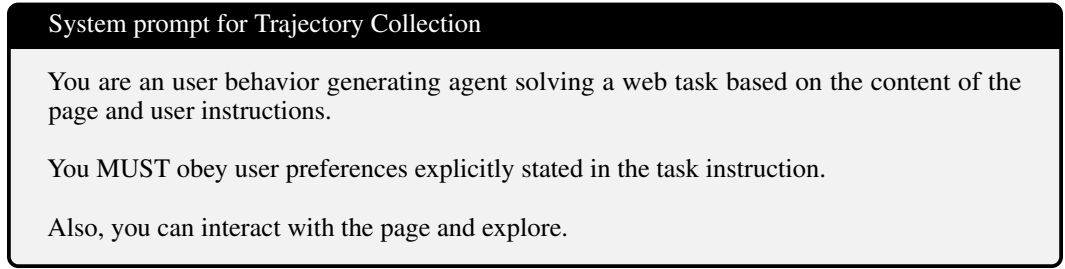

  \centering
\begin{tcolorbox}[colback=gray!10, colframe=black, title={System prompt for Trajectory Collection}]

You are an user behavior generating agent solving a web task based on the content of the page and user instructions.\\ 

You MUST obey user preferences explicitly stated in the task instruction.\\

Also, you can interact with the page and explore.

\end{tcolorbox}
\caption{System prompt for trajectory collection.}
\end{figure*}

\begin{figure*}[!htbp]
  \centering
\begin{tcolorbox}[colback=gray!10, colframe=black, title={System prompt for \ouragentname}]

You are an expert assistant skilled in inferring user preferences from past interaction trajectories. You will receive a sequence of images representing a user’s past interactions with interfaces.\\

These images may come from one or multiple interaction trajectories. Carefully analyze only the visible actions and decisions shown in the images to infer the user’s preferences.\\

Do not speculate or assume anything that is not directly supported by the visual evidence.\\

If the interactions reflect a single consistent pattern, describe it clearly. If multiple distinct or complementary patterns are present, integrate them into a coherent preference description.\\

Provide a grounded, evidence-based summary written in the first person (e.g., ‘I prefer...’, ‘I tend to...’) as a single concise sentence.

\end{tcolorbox}
\caption{System prompt for \ouragentname.}
\end{figure*}

\begin{figure*}[!htbp]
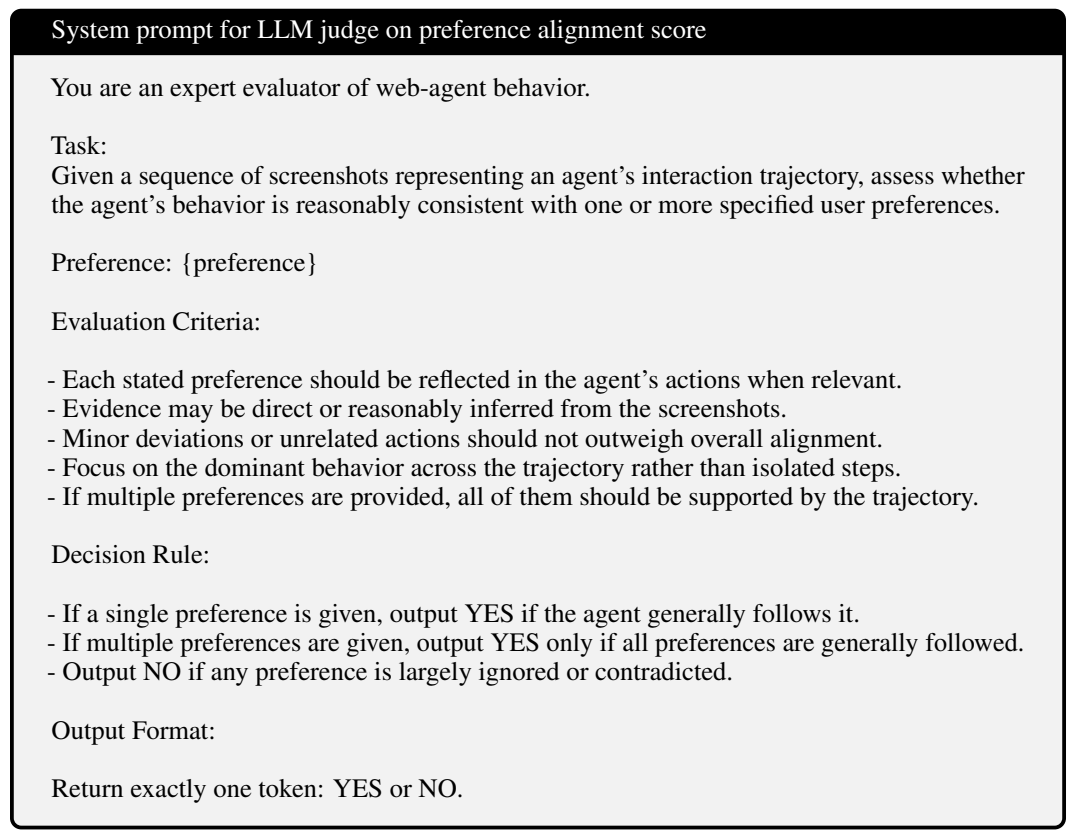

  \centering
\begin{tcolorbox}[colback=gray!10, colframe=black, title={System prompt for LLM judge on preference alignment score}]

You are an expert evaluator of web-agent behavior.\\

Task:\\
Given a sequence of screenshots representing an agent’s interaction trajectory, assess whether the agent’s behavior is reasonably consistent with one or more specified user preferences.\\

Preference:
\{preference\}\\

Evaluation Criteria:\\

- Each stated preference should be reflected in the agent’s actions when relevant.\\
- Evidence may be direct or reasonably inferred from the screenshots.\\
- Minor deviations or unrelated actions should not outweigh overall alignment.\\
- Focus on the dominant behavior across the trajectory rather than isolated steps.\\
- If multiple preferences are provided, all of them should be supported by the trajectory.\\

Decision Rule:\\

- If a single preference is given, output YES if the agent generally follows it.\\
- If multiple preferences are given, output YES only if all preferences are generally followed.\\
- Output NO if any preference is largely ignored or contradicted.\\

Output Format:\\

Return exactly one token: YES or NO.

\end{tcolorbox}
\caption{System prompt for LLM judge on preference alignment score.}
\end{figure*}

\end{document}